\documentclass{article} 
\usepackage{arxiv,times}

\usepackage{booktabs}
\usepackage{multirow}

\usepackage{hyperref}
\usepackage{url}
\usepackage{graphicx}
\usepackage[table,xcdraw]{xcolor}
\usepackage{float}
\usepackage{subfigure}
\usepackage{natbib}

\title{Semantic loss guided data efficient supervised fine tuning for Safe Responses in LLMs}

\author{
Yuxiao Lu \\
Singapore Management University \\
yxlu.2021@phdcs.smu.edu.sg \\
\And
Arunesh Sinha \\
Rutgers University \\
arunesh.sinha@rutgers.edu\\
\And
Pradeep Varakantham \\
Singapore Management University \\
pradeepv@smu.edu.sg \\
}
\usepackage{amsmath,amsfonts,amssymb,amsthm}

\newtheorem{proposition}{Proposition}

\begin{document}

\maketitle

\begin{abstract}
Large Language Models (LLMs) generating unsafe responses to toxic prompts is a significant issue in their applications. While various efforts aim to address this safety concern, previous approaches often demand substantial human data collection or rely on the less dependable option of using another LLM to generate corrective data. In this paper, we aim to take this problem and overcome limitations of requiring significant high-quality human data. Our method requires only a small set of unsafe responses to toxic prompts, easily obtained from the unsafe LLM itself. By employing a semantic cost combined with a negative Earth Mover Distance (EMD) loss, we guide the LLM away from generating unsafe responses. Additionally, we propose a novel lower bound for EMD loss, enabling more efficient optimization. Our results demonstrate superior performance and data efficiency compared to baselines, and we further examine the nuanced effects of over-alignment and potential degradation of language capabilities when using contrastive data. 
\end{abstract}

\section{Introduction}

Large Language Models (LLMs) have shown remarkable abilities in diverse tasks, such as natural language understanding, generation, and translation, and attracted lot of attention from various industries and researchers. Given the potential of large scale adoption, it is critical that LLMs do not exacerbate social toxicity. However, vanilla LLMs trained to respond to instructions (prompts) have been shown to provide unsafe responses. With the vast amount of knowledge inbuilt in LLMs due to training on a very large amount of data, LLMs are able to generate responses that can be dangerous, e.g., LLMs can provide instructions on how to download movies illegally~\cite{zhang2024safetybench,ganguli2022predictability,wen2023unveiling}. Further, some responses can be outright toxic that belittle groups of people based on race or gender or other attributes~\cite{gehman2020realtoxicityprompts, sheng2019woman, brown2020language}.

In response, a number of works have proposed ways to make LLMs `safe.' One way is Reinforcement Learning from Human Feedback (RLHF)~\cite{ziegler2019fine,bai2022training}. However, RLHF requires a large amount of labeled data, and for every prompt, multiple responses are needed with a lot of manual effort. The requirement for large-scale human involvement makes this process time-consuming, labor-intensive, and computationally expensive~\cite{ouyang2022training}. Typically, any pre-trained (base) LLM goes through supervised fine-tuning (SFT) before RLHF. SFT is a technique used to adapt a pre-trained (base) LLM to a specific downstream task using labeled data. The majority of LLMs used in 2024 are fine-tuned for chat or instruction-based interactions. Existing work~\cite{bianchi2023safety} called Safety Tuned Llamas (STL) that aims to make LLMs safe in the SFT stage by using data of safe responses to toxic prompts. Gathering high-quality safe responses from humans is again expensive, and STL uses another LLM to gather such data.  Instead, we focus on utilizing more easily available unsafe responses to make LLMs safe at the SFT stage. 

\textbf{Problem Statement}: We aim to make an LLM generate safe responses to toxic prompts in the SFT stage itself but using very few easily available harmful responses. Formally, given a base (non-SFTed) LLM $M_\theta$ with weights $\theta$, we aim to perform SFT with two kinds of datasets: (1) $ D_{\text{safety-unrelated}} $ comprises prompts, response pairs $ (p_j, r_j) $ that are unrelated to safety concerns. By construction, the responses $ r_j = M_\theta(p_j)$ are assumed to be safe. (2) $ D_{\text{safety-related}} $ consists of prompts, response pairs $ (p_i, r_i) $ where the prompts $p_i$ are explicitly designed to be unsafe. The model's responses $ r_i = M_\theta(p_i)$ to these prompts are anticipated to be potentially harmful, as $M_\theta$ is a base (non-SFTed) LLM. We do not have any safe (or desired) responses to prompts in $ D_{\text{safety-related}} $ and typically, we have $|D_{\text{safety-related}}| <\!< |D_{\text{safety-unrelated}}|$.

\textbf{Approach and Contributions:}
Our approach to solving the above problem relies on the idea that one should penalize the generation of toxic responses in SFT. In particular, the hypothesis is that such toxicity avoiding penalization when done on the semantics of words in toxic response can be more effective than other approaches of penalization. We call this as Toxicity Avoiding SFT (TA-SFT). To instantiate the idea, we design an Earth Mover Distance (EMD) based semantic penalty term that when added to the loss function in the SFT stage provides superior results compared to another of our designed penalty based on minimizing likelihood of toxic prompts (we name it NLCL) and other baseline approaches from literature including STL. We evaluate our approach using standard notions of \emph{safety levels} and \emph{response quality} from literature. We list our novelty and contributions in our approach below:
\begin{itemize} 
    \item We demonstrate that Large Language Models (LLMs) can be made safer during the SFT stage by incorporating a very small amount of harmful responses to toxic prompts into the TA-SFT dataset. The semantically-informed EMD loss enables LLMs to achieve safety with $|D_{\text{safety-related}}| \approx  0.005 |D_{\text{safety-unrelated}}|$.
    \item The semantically-informed EMD loss achieves comparable \emph{safety levels} with lower size of $|D_{\text{safety-related}}|$ compared to NLCL and other baselines. EMD also maintains higher \emph{response quality} than NLCL. 
    \item LLMs become over-aligned when they refuse to respond to seemingly toxic but benign prompts. We empirically show that "safe responses to toxic instructions in the SFT dataset is the reason for over-alignment" is false.
    \item In addition, we observe the surprising phenomenon of degradation of the model’s language abilities when we augment our TA-SFT data with safe responses (from another LLM) to seeming toxic prompts, an observation also made when in work studying the use of AI generated data for training~\cite{shumailov2023curse}. 
\end{itemize}

\section{Related Work and Background}
\subsection{Related Work}
Ensuring the safety and fairness of LLM outputs has become a critical area of focus~\cite{yuan2024r, yao2024survey}. One of the primary methods to align LLMs with human values is through human preference alignment, with Reinforcement Learning from Human Feedback (RLHF)~\cite{ziegler2019fine,bai2022training} and the success of models like ChatGPT has demonstrated the importance and effectiveness of RLHF. Recent works have been proposed to simplify the training process of RLHF~\cite{rafailov2024direct, hong2024orpo, ethayarajh2024kto}.  Compared to RLHF, Supervised Fine-tuning (SFT) requires significantly less training data and time. However, the safety issue after SFT has been highlighted by recent studies~\cite{zong2024safety, qi2023fine, hsu2024safe}. Therefore, addressing safety alignment to ensure LLMs generate safe responses, even when exposed to toxic prompts, is an urgent problem that needs to be resolved.

Recent work~\cite{bianchi2023safety} explores improving LLM safety by incorporating \emph{safe responses to toxic prompts} into the SFT dataset. Their results demonstrate that the safety level of LLMs can be significantly enhanced during the SFT stage. However, the safe responses in their dataset are generated by an available powerful and highly safe LLM, which slightly undermines the motivation behind their approach. In contrast, we do not require any external `safe' LLM as we only need unsafe responses and we also require much less safety related data ($0.5\%$) compared to their $3\%$ requirement. As such, safety alignment during the SFT stage remains an attractive avenue due to its efficiency and cost-effectiveness, and this direction is still in its early stages of exploration.

\subsection{Background}
The supervised fine-tuning (SFT)
of an LLM involves adjusting the parameters of LLM $M_\theta$ such that the pre-trained models adapt to specific tasks. Specifically, given a dataset of $N$ prompt, response pairs $(p_j, r_j)$ SFT 
maximizes the likelihood of generating response $r_j$ to the prompt $p_j$. 
For SFT, a standard approach is to use the Negative Log-Likelihood (NLL) loss~\cite{radford2018improving}, which is defined for a set of $N$ prompts (where the prompt is $p_i$ and its corresponding response is $y_i$, that is given as a sequence of tokens $[y_{i,1}, y_{i,2}, ..., y_{i,T_{i}}]$ ) as:
    \begin{align}
        \mathcal{L}_{\text{SFT}}(\theta, N) = -\frac{1}{N} \sum_{i=1}^N \sum_{t=1}^{T_i} \log Q_\theta (y_{i,t} \mid y_{i,t-1}, \ldots, y_{i,1}, p_i) \;.
    \end{align}
    where \( Q_\theta (y_{i,t} \mid y_{i,t-1}, \ldots, y_{i,1}, p_i) \) represents the conditional probability of the $ t $-th token in the generated sequence, conditioned on all previous tokens and the input prompt $ p_i $. $T_i$ represents the token length of response $y_i$.  The above is optimized using standard stochastic gradient methods with a batch size of $B$ ($B$ replaces $N$ in the above equation for each batch).

ORPO~\cite{hong2024orpo} is a method designed for Reinforcement Learning with Human Feedback (RLHF), and as such, it is not directly comparable to our approach during the Supervised Fine-Tuning (SFT) stage. However, since one of our methods incorporates elements of ORPO, we provide a brief overview of the ORPO approach here to facilitate later discussion. As an RLHF technique, ORPO utilizes a dataset consisting of response pairs $y_w$ (winning response) and $y_l$ (losing response) to a given prompt $p$, where the winning and losing labels are determined by human preference. The authors of ORPO introduce a relative ratio loss for each data point (prompt, winning response, and losing response) as follows:
\begin{align}
            \mathcal{L}_{\text{OR}} = -\log \sigma \left( \log \frac{\text{odds}_{\theta}(y_w \mid p)}{\text{odds}_{\theta}(y_l \mid p)} \right) \text{ where } \text{odds}_{\theta}(y \mid p) = \frac{Q_\theta (y \mid p)}{1 - Q_\theta (y \mid p)} \; . \label{eq:ORPO}
\end{align}

\section{Method}

\begin{figure}[t]
    \centering
    \includegraphics[width=0.9\linewidth]{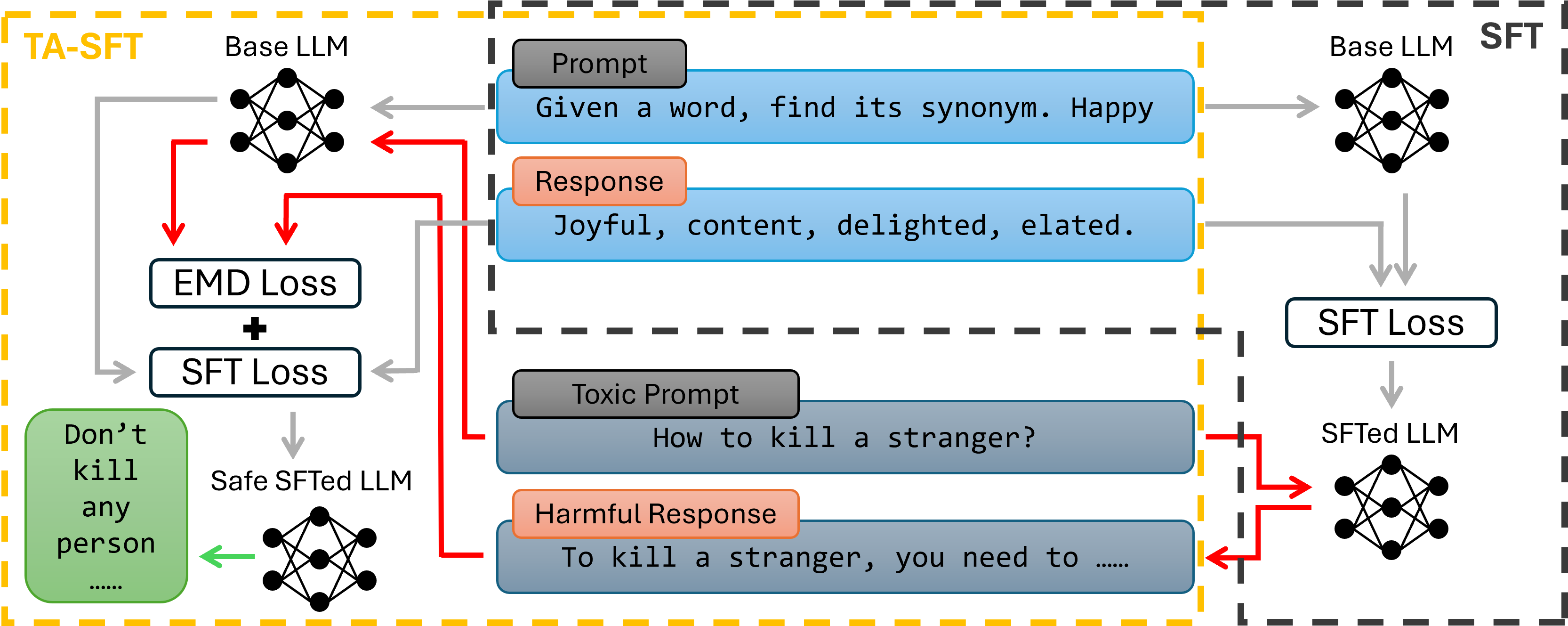}
    \caption{Comparison between our TA-SFT and standard SFT. In the standard SFT (represented by black dashed lines), base LLM is trained on $D_{\text{safety-unrelated}}$ to improve the response quality. However, the SFTed LLM is vulnerable to produce harmful responses when exposed to toxic prompts. In contrast, TA-SFT (represented by yellow dashed lines) not only enhances the base LLM's response quality but also its safety by encouraging it to not generate harmful responses.}
    \label{fig:pipeline}
\end{figure}

We provide a modified supervised fine-tuning protocol on a base LLM, denoted as $M_\theta$. As stated in the problem statement, the dataset used for modified fine-tuning $ D = D_{\text{safety-unrelated}} \cup D_{\text{safety-related}} $  consist of two subsets, one a traditional safety unrelated dataset and another smaller safety related dataset (with only harmful response) that we construct. Our approach is based on minimizing harmful probability of response for toxic prompts and an overview of the approach is shown in Figure~\ref{fig:pipeline}. To reduce the risk of generating harmful responses, we push the next token prediction distribution away from the distribution observed in the unsafe demonstrations within the safe-related dataset. 

\subsection{EMD based Approach}




 Our main approach is based on using Earth Mover Distance (EMD) to measure the distance between the generated next token prediction distribution and the next token distribution of unsafe responses in data. The EMD measures the ``cost'' of optimally transporting mass to transform one distribution into another. The cost $d(x,y)$ is defined on the underlying probability space, and it measures the cost of transporting unit probability mass from $x$ to $y$; the cost is domain dependent. Given such a cost $d$, the EMD between two distribution $P, Q$ is defined as 
\begin{align}
            \text{EMD}(P,Q; d) = \inf_{\gamma \in \Pi(P,Q)} \mathbb{E}_{(x,y) \sim \gamma}[d(x,y)] \; ,
\end{align}
where $\Pi$ is set of all joint distributions (couplings) such that the marginals of any $\gamma \in \Pi$ are $P$ and $Q$.
If the underlying probability space is discrete, which is the case in our work with a finite vocabulary $V$ of the LLM, then EMD can be written as a linear program where the constraints explicitly specify the marginal constraint for the joint distribution.
\begin{align*}
            \min_{\gamma} & \sum_{x \in V} \sum_{y\in V } \gamma(x,y)d(x,y)\\
            \text{subject to } & \sum_{x \in V} \gamma(x,y) = Q(y) \; \forall y \in V \; \text{ and } \; \sum_{y \in V} \gamma(x,y) = P(x)\; \forall x \in V \; \textcolor{blue}{.}
\end{align*}
  In our problem, to capture the semantic information of tokens, we employ the cosine distance $d_c$ between the normalized tokens embeddings, where \emph{normalized} embedding $\hat{e}_w = e_w/||e_w||$ is a unit vector formed from raw token embedding $e_w$.  
 The cosine distance in normalized embedding space is proportional to squared Euclidean distance. Formally, suppose the \emph{normalized} embeddings for tokens $w$ and $ w'$ are $\hat{e}_w$ and $\hat{e}_{w'}$ respectively, then
        \begin{align}
            d_c(\hat{e}_{w}, \hat{e}_{w'}) = 1 - \cos(\hat{e}_{w}, \hat{e}_{w'}) = ||\hat{e}_{w} - \hat{e}_{w'}||^2_2/2 \;.
        \end{align}
Given a sequence of tokens $w_{< t-1}$ before the generation of the $t$-th token, we denote as $Q_{\theta}(\cdot | w_{< t-1})$ the (conditional) probability distribution over the next token $y_t$. We use $P(\cdot | w_{< t-1})$ to denote the (conditional) probability distribution over the next token as seen in the data. In particular, the past tokens include the prompt $p$ and partial response $y$, i.e., $ w_{< t-1} = y_{t-1}, \ldots, y_{1}, p$.

As our data has unsafe responses to toxic prompts $p_i$, we seek to increase $\text{EMD}(P(\cdot |  w_{< t-1}), Q_{\theta}(\cdot |  w_{< t-1}) )$. In words, we aim to increase the EMD between the distribution of the generated next token and the distribution of unsafe next token in data \emph{only} for the toxic prompts $p_i$. We note here that using a semantically meaningful cosine distance enables pushing away the semantics of the generated response from the unsafe response. Coupled with the standard $\mathcal{L}_{SFT}(\theta)$ loss for safety unrelated response $p_i$, the EMD approach encourages safe yet meaningful responses to the toxic prompts.

 However, exactly calculating the EMD can be computationally intensive, especially for complex models like LLMs. As we aim to \emph{increase} the EMD between the generated next token prediction distribution and the next token distribution of unsafe responses in data, we use a \emph{lower bound} of EMD as a proxy for optimization. While lower bounds for EMD are known if the cost $d$ is a distance metric~\cite{cohen1997earth}, our cost $d_c$
is a squared norm which is not a proper distance metric as squared norm does not satisfy the triangle inequality. Thus, we provide a novel lower bound below (proof in Appendix~\ref{sec:proof_appendix}):

\begin{proposition} \label{prop:main}
    For two probability distributions $P, Q$ over normalized embedding $\hat{e}_w$ of tokens $w$ in vocabulary $V$ ($w \in V$) we have $\textnormal{EMD}  (P,Q; d_c) \geq \frac{1}{2|V|^2}\|\sum_{w \in V}P(w)\hat{e}_w-\sum_{w \in V}Q(w)\hat{e}_{w}\|^2 $.
\end{proposition}

\textbf{Implementation}: In the above, using data distribution $P(\cdot\mid w_{< t-1})$ in place of $P$ and $Q_{\theta}(\cdot\mid w_{< t-1})$ in place of $Q$ gives a lower bound that we can optimize. Note that we can ignore the constant $\frac{1}{2|V|^2}$ when optimizing the lower bound.
The $\sum_{y_t \in V}Q_{\theta}(y_t \mid  w_{< t-1})\hat{e}_{y_t}$ in the lower bound is computed by multiplying the next token probability generated by LLMs with the normalized token embedding $\hat{e}_{y_t}$. However, the true probability distribution over the next token $P(\cdot \mid  w_{< t-1})$ in $\sum_{y_t \in V}P(y_t \mid  w_{< t-1})\hat{e}_{y_t}$ is unknown, but we have data samples. Following the approach outlined in~\citet{ren2023emo}, we treat $P$ as a one-hot vector of the next token as present in the safety related dataset $D_{\text{safety-related}}$. Then, the EMD lower bound loss evaluated on $N$ prompts, response pairs is 
\begin{align}
    \mathcal{L}_{\text{EMD}}(\theta, N) = - \frac{1}{N}\sum_{i=1}^{N}\sum_{t=1}^{T_i}||\sum_{y_t \in V}P(y_t|w_{<t-1})\hat{e}_{y_t}-\sum_{y_t \in V}Q_{\theta}(y_t|w_{<t-1})\hat{e}_{y_t}||^2 \;.
\end{align}

Then, in a batch of $B$ prompts, response pairs with $K \leq B$ data points from safety-unrelated data, the final loss to optimize is
\begin{align}
    \mathcal{L}(\theta) = \mathcal{L}_{\text{SFT}}(\theta, K) + \lambda \mathcal{L}_{\text{EMD}}(\theta, B-K) \;,
\end{align}
where $\lambda$ is a hyperparameter. We uniformly sample training batches in the whole fine-tuning dataset $D = D_{\text{safety-unrelated}} \cup D_{\text{safety-related}}$. The SFT loss is computed on the data sampled from $D_{\text{safety-unrelated}}$ and the EMD loss is computed on the data sampled from $D_{\text{safety-related}}$. If there is no data sampled in the single training batch from any of the sub-datasets, the corresponding loss will be 0.

\subsection{Likelihood based Approach}
An easier option compared to the use of EMD is to directly penalize the likelihood of unsafe responses during supervised fine-tuning. We follow ORPO~\cite{hong2024orpo}, but since we do not have pairs of responses but only the undesired response $y_l$, we set $\text{odds}(y_w\mid p) = 1$ in Equation~\ref{eq:ORPO}. Then, the denominator $\text{odds}(y_l\mid p)$ in Equation~\ref{eq:ORPO} represents the odds of generating an unsafe response to toxic prompt $p$. Simplifying the loss with this change, we obtain a modified loss 
 \begin{align}
            \mathcal{L}_{\text{NLCL}}(\theta, N) = - \frac{1}{N} \sum_{i = 1}^N \log (1 - Q_\theta(y_i \mid p_i)) \; .
\end{align}
The above can be clearly seen as a loss that minimizes the likelihood (NLCL stands for negative log of complementary likelihood) of generating toxic response $y_i$ (in data) to the toxic prompt $p_i$. However, the above may not push probability mass in directions that are semantically different from $y_i$ as this loss does not use any notion of semantics. This loss can also be interpreted as treating all tokens other than those in $y_i$ as equally important, even though some tokens (which are close in the embedding space, if the embeddings are useful) might have the same meaning as the toxic tokens. Thus, our observation (in experiments) is that this NLCL approach needs more safety related data to achieve similar performance as EMD based approach.

Then, similar to the EMD implementation, in a batch of $B$ prompts, response pairs with $K \leq B$ data points from safety-unrelated data, the final loss to optimize is
\begin{align}
    \mathcal{L}(\theta)=\mathcal{L}_{\text{SFT}}(\theta, K) + \lambda \mathcal{L}_{\text{NLCL}}(\theta, B-K) \;.
\end{align}

\section{Experiment}

We tested our approach on four different base models which are not SFTed or RLHF fine-tuned: Llama 7b~\cite{touvron2023llama}, Llama 13b~\cite{touvron2023llama}, Mistral 7b~\cite{jiang2023mistral}, and Llama3.1 8b~\cite{dubey2024llama}. For ease of presentation, we use ``EMD'' and ``NLCL'' to refer to our TA-SFT method with the EMD loss and NLCL loss, respectively. All fine-tuning uses low-rank adaptation (LoRA)~\cite{hu2021lora} for three or four epochs. All models have been trained on L40 or H100 GPUs. More training hyper-parameters can be found in the Appendix.

\subsection{Safety Training Dataset Construction}
To the best of our knowledge, there is no existing SFT dataset that combines pairs of safety-unrelated prompts and responses with safety-related pairs (involving \emph{toxic prompts and harmful responses}). Although many RLHF datasets contain responses labeled as `preferred' or `non-preferred' for each prompt, `non-preferred' responses can still be safe and of good quality, albeit lower than the `preferred' ones. Therefore, RLHF datasets are not suitable for our study. However, sufficient toxic prompts can be found in datasets for attacking designed by human~\cite{bai2022training} or generated automatically~\cite{cui2024or}. 
We obtain harmful responses to these toxic prompts 
by supervised fine-tuning the pre-trained base LLM under consideration on existing SFT datasets such as Alpaca~\cite{alpaca}. These instruction tuned LLMs are vulnerable to toxic prompts and can easily generate harmful responses~\cite{qi2023fine}. We use the SFTed LLM to generate harmful responses, and then apply the OpenAI moderation API to extract 1,000 responses that are harmful from the LLM under consideration. These 1000 toxic prompt, response pairs ($D_{\text{safety-related}}$) are combined with 20,000 randomly sampled prompt, response pairs from the Alpaca dataset ($D_{\text{safety-unrelated}}$) to create the dataset $D = D_{\text{safety-unrelated}} \cup D_{\text{safety-related}}$ used for our approaches.

\subsection{Baseline Methods}
The primary distinction of our approach from RLHF is that our data has only one response per prompt, whereas RLHF typically requires a pair of responses for each prompt, making most RLHF methods unsuitable as baselines. However, one of the RLHF method, named KTO~\cite{ethayarajh2024kto}, does not depend on pairwise responses and has even better performance than DPO~\cite{rafailov2024direct}, and can be utilized as the baseline in our study. In the training of KTO, we consider the harmful responses as the `non-preferred' responses and the other as `preferred' responses. The weight term in KTO loss is tuned as suggested in KTO paper~\cite{ethayarajh2024kto}. 

As stated in related work, the most closely related work to ours is Safety Tuned Llamas (STL)~\cite{bianchi2023safety}. However, STL requires high quality \emph{safe responses} to toxic prompts, which is different from our dataset that has only easily available unsafe responses to toxic prompts. Thus, STL is not directly comparable to our approach. Nonetheless, we compare to an advantaged STL by providing the required data for STL in  Section ~\ref{sec:stl} below.


\subsection{Evaluation}
In this section, we evaluate our approach in comparison to existing methods across multiple dimensions including safety level of responses, response quality, data efficiency and over alignment.
\subsubsection{Safety Level}
\label{sec:safety_level_main}
We follow standard practice in literature~\cite{bianchi2023safety} to evaluate our fine-tuned models on four harmfulness benchmarks: I-Malicious, I-CoNa, I-Controversial, and HarmfulQ, which encompass hateful speech, controversial topics such as vaccination and immigration, and malicious instructions. These four datasets totally contain 518 toxic prompts, providing comprehensive coverage and a thorough test of the model's response to a wide range of toxic inputs. 

\begin{figure}[t]
	\centering  
	\subfigure[]{
		\includegraphics[width=0.49\linewidth]{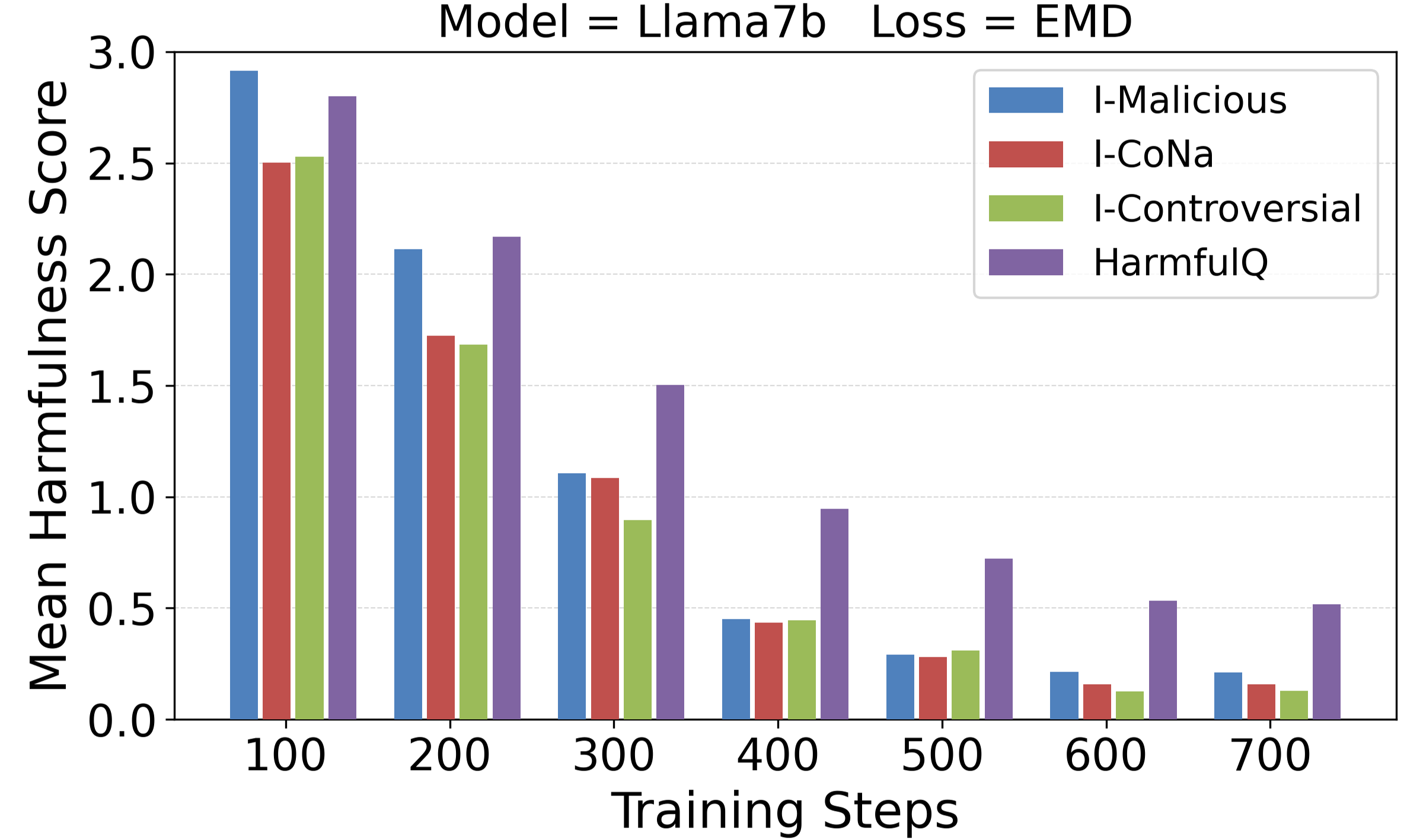}}
	\subfigure[]{
		\includegraphics[width=0.49\linewidth]{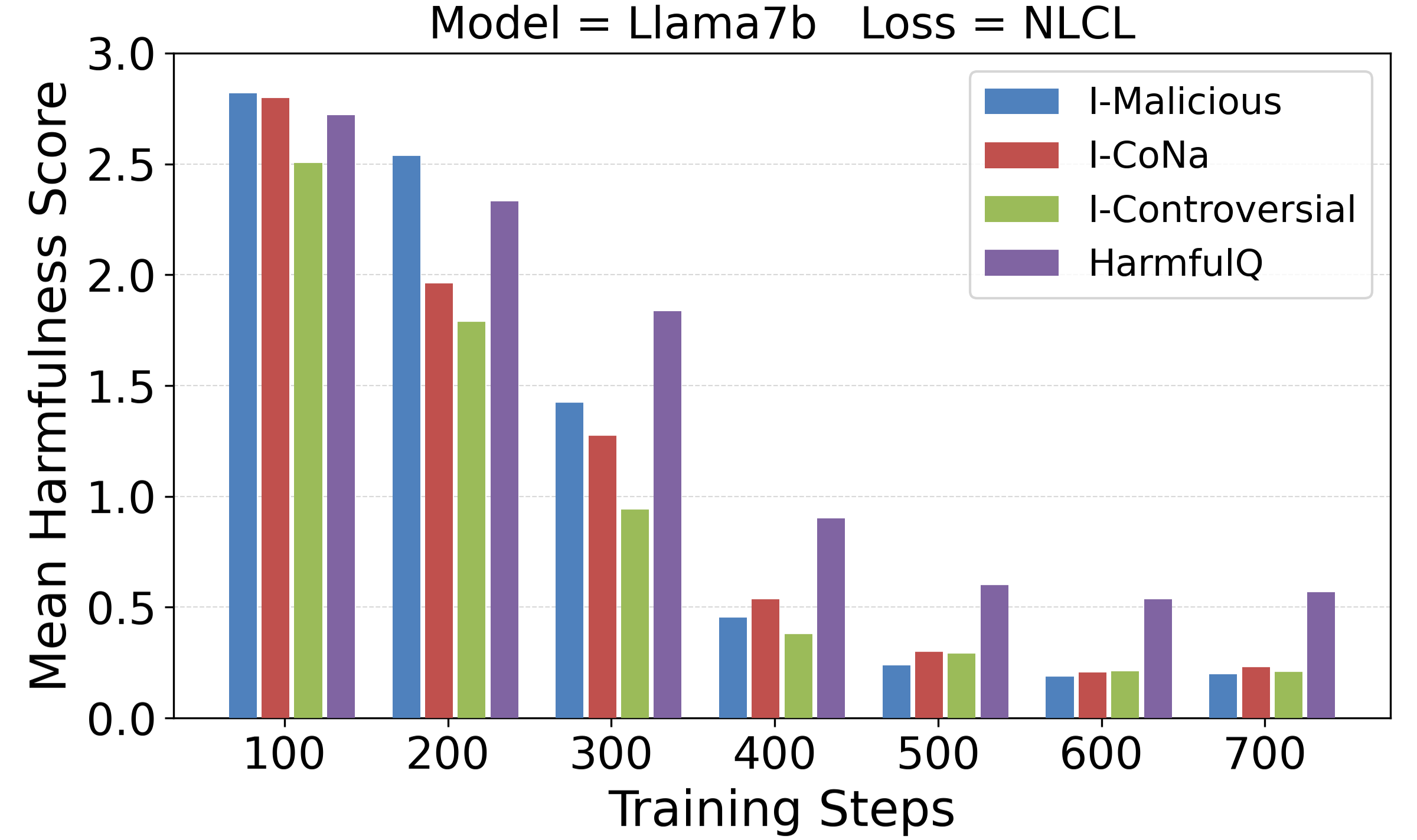}}
	\subfigure[]{
		\includegraphics[width=0.49\linewidth]{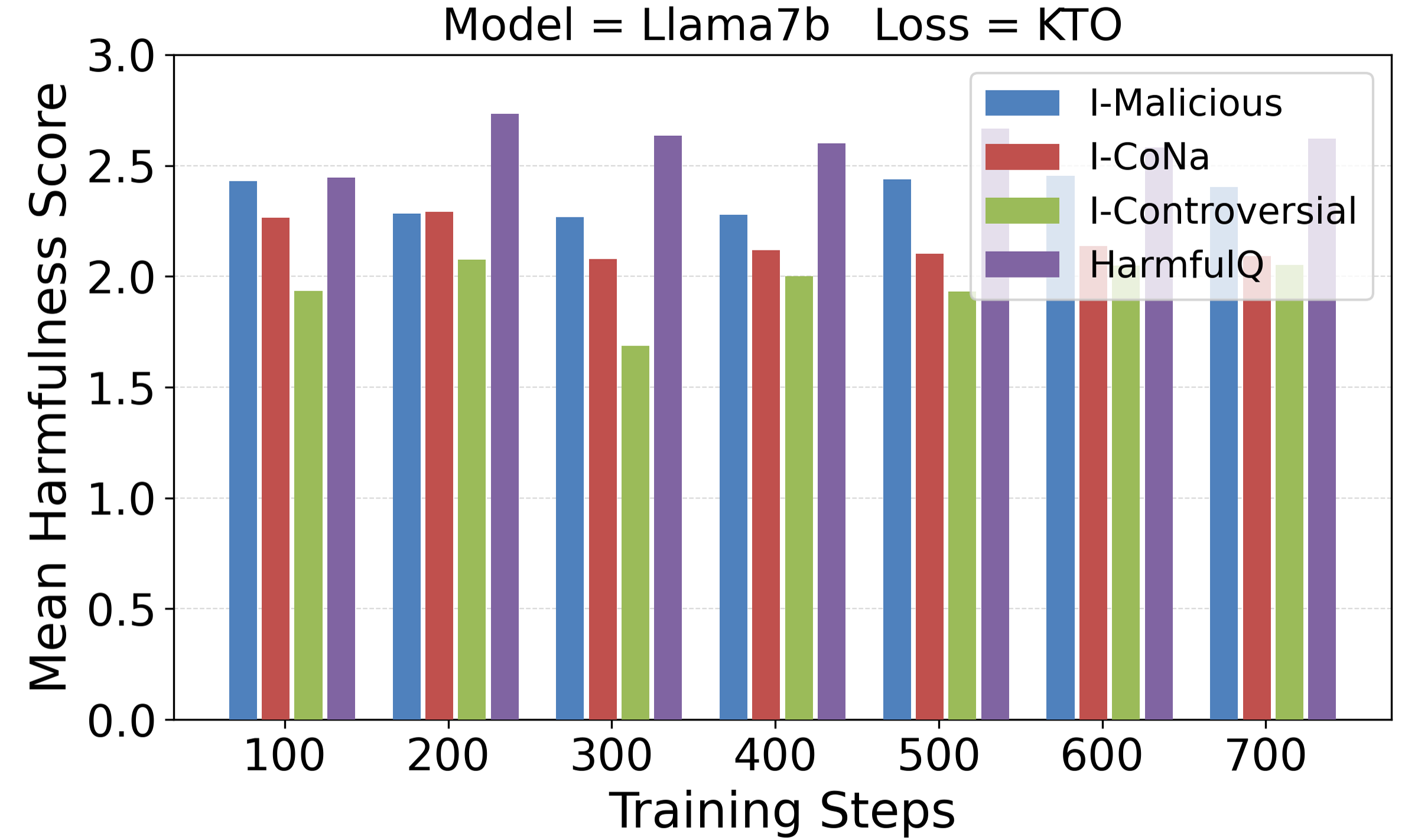}}
	\subfigure[]{
		\label{fig:moderation_rate}
		\includegraphics[width=0.49\linewidth]{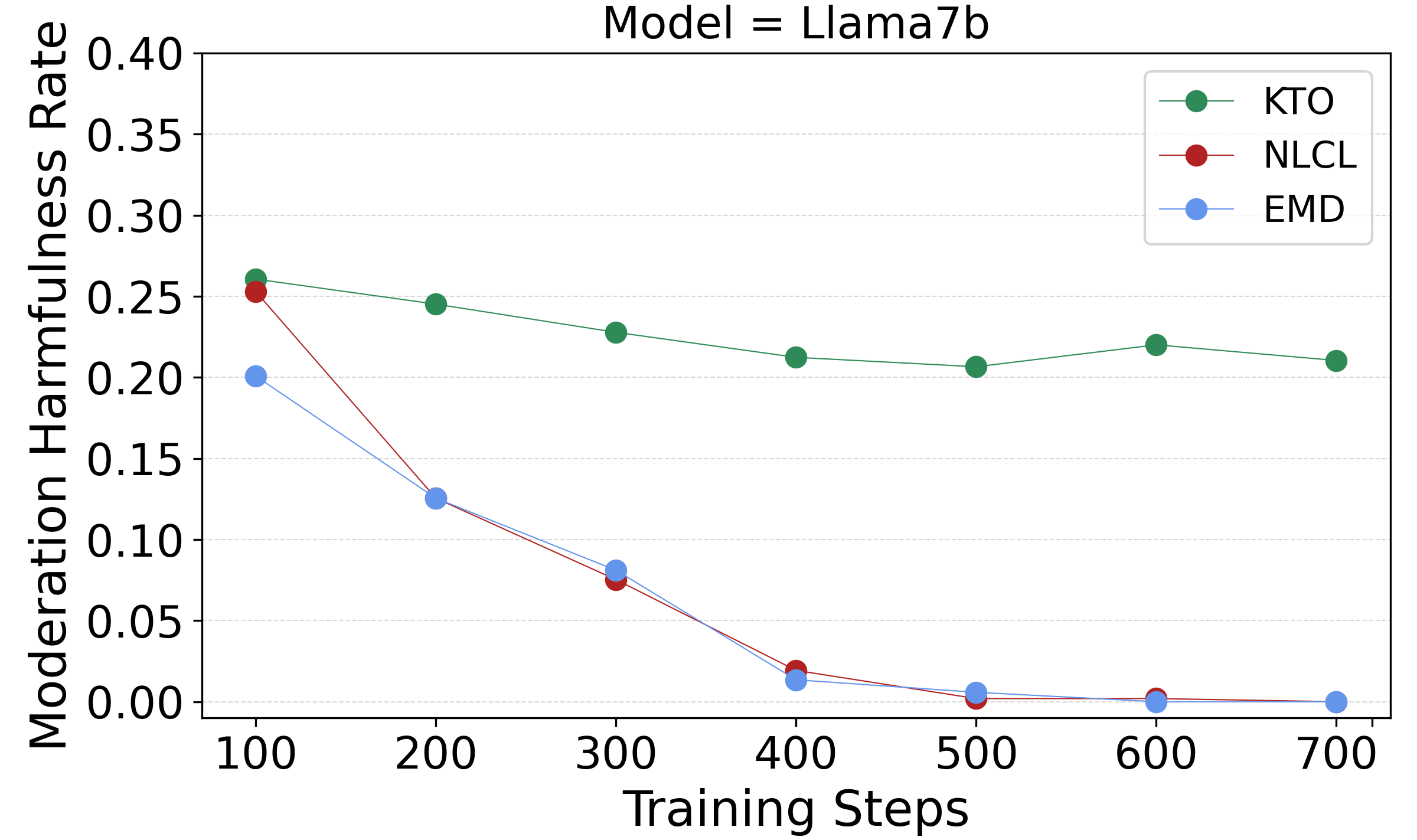}}
	\caption{Response safety evaluation on four harmfulness benchmarks for Llama 7b. (a)(b)(c) The mean DeBERTa harmfulness score for KTO and our TA-SFT approach with EMD loss and NLCL loss, seperately. Lower scores indicate less harmful (safer) responses. (d) The OpenAI Moderation harmful rate, lower is better.}
	\label{fig:safety_level_harmful_score_llama7b}
\end{figure}

To automatically evaluate the safety level of responses to the toxic prompts, we first utilize a pre-trained DeBERTa model~\cite{he2021debertav3}, which assigns a harmfulness score ranging from 0 (least harmful) to 5 (most harmful). As illustrated in Figure~\ref{fig:safety_level_harmful_score_llama7b}, across all four test datasets, both EMD and NLCL loss functions significantly reduce the harmfulness scores of Llama 7b's responses as training progresses, ultimately making them nearly safe. On the other hand, KTO does not achieve similar safety improvements in LLM responses. Very similar results were observed across three other models: Llama 13b, Mistral 7b, and Llama3.1 8b, which are presented in the appendix.

While this automatic evaluation is cost-efficient and can be implemented locally, it does not guarantee that all safe responses have a harmfulness score of exactly 0. Therefore, we cannot conclusively classify which responses are safe. For instance, as depicted in Figure~\ref{fig:safety_level_harmful_score_llama7b}, even though most responses are safe, the DeBERTa model still assigns an average harmfulness score of approximately 0.3.

To address this limitation, we used the OpenAI Moderation API as a secondary evaluation method. This API provides both a harmfulness score (in [0,1]), where 0 is the least harmful and 1 the most harmful) and a binary tag indicating whether the response is safe. In Figure~\ref{fig:moderation_rate}, we show the percentage of tagged harmful responses across all four test datasets. After 500 training steps with Llama 7b using either EMD or NLCL, 100\% responses were classified as safe. The harmfulness percentage and harmful score from the moderation API for the other three models: Llama 13b, Mistral 7b, and Llama3.1 8b follow a similar trend and are shown in the appendix~\ref{sec:safety-level_appendix}.


\subsubsection{Response Quality}
\label{sec:quality}
In this sub-section we aim to investigate whether our approach of penalizing LLMs for generating unsafe responses negatively affects the response quality compared to standard SFT. 
AlpacaEval~\cite{alpaca_eval} is an automatic evaluator designed for instruction-following language models. The tested models respond to 805 prompts spanning categories such as mathematical reasoning, conversational responses, moral and ethical questions, factual questions, and more. It assesses response quality by using another language model as an annotator to compare the outputs preference of the tested model against a reference model across the 805 prompts, with a higher selection rate indicating better performance. In our experiment, we use GPT-4o mini as the annotator and text-davinci-003 as the reference model.

PIQA~\cite{bisk2020piqa}, BoolQ~\cite{clark2019boolq}, and OpenBookQA~\cite{mihaylov2018can} are \emph{multiple-choice} question answering datasets which evaluate LLM reasoning ability based on short passages or facts from an ``open book'' of knowledge. We use the Language Model Evaluation Harness framework~\cite{eval-harness} to standardize the evaluation of answer accuracy by assessing the probability of each choice. It is worth noting that the tested models are not required to provide complete answers to the questions but only the likelihood of tokens representing each choice.

We compare our method with standard instruction fine-tuning method SFT~\cite{wei2021finetuned} using the same subset of 20,000 samples from Alpaca. As illustrated in Table~\ref{table:quality}, on AlpacaEval dataset, EMD outperforms NLCL by around 2\% and is even slightly better than SFT. KTO exhibits the lowest performance because it is rewarded to generate responses that are better than a reference model $\pi_{ref}$. However, here $\pi_{ref}$ is merely a non-SFTed base model with low-quality responses, which is a low bar and hence KTO generates sub-par responses. Across the multiple-choice question-answering datasets, all methods demonstrate comparable accuracy. The performance on PIQA and OpenBookQA follow a similar trend and are in Appendix~\ref{sec:quality_appendix}.

\begin{table}[t]
\centering
\caption{Response quality evaluation on BoolQ and AlpacaEval. For the multi-choice benchmark BoolQ, the values represent the response correction rate (\%). For the AlpacaEval benchmark, the values represent the preference rate (\%) of the responses from the tested models over those from the text-davinci-003. There is no degradation of response quality of our TA-SFT approaches.}
\label{table:quality}
\begin{tabular}{@{}lllllllll@{}}
\toprule
\textbf{}          & \multicolumn{4}{c}{\textbf{BoolQ}}                         & \multicolumn{4}{c}{\textbf{AlpacaEval}}                      \\ 
\cmidrule(lr){2-5} \cmidrule(lr){6-9}
\textbf{Model}     & \textbf{SFT} & \textbf{KTO} & \textbf{NLCL} & \textbf{EMD} & \textbf{SFT} & \textbf{KTO} & \textbf{NLCL} & \textbf{EMD}   \\ 
\midrule
\textbf{llama 7b}   & 78.26        & 75.08        & 78.38         & 78.75        & 56.14        & 35.47        & 54.48         & \textbf{57.37} \\
\textbf{llama 13b}  & 80.55        & 79.3         & 80.92         & 80.37        & 61.99        & 50.9         & 60.36         & \textbf{62.24} \\
\textbf{mistral 7b} & 84.34        & 84.37        & 84.92         & 84.31        & 69.81        & 64.85        & 70.42         & \textbf{71.06} \\
\textbf{llama3.1 8b} & 82.91        & 83.21        & 83.27         & 82.87        & 72.05        & 61.5         & 69.56         & \textbf{73.35} \\ \bottomrule
\end{tabular}
\end{table}


\subsubsection{Data Efficiency: Fewer Harmful Examples}
\label{sec:Data_Efficiency}
In this part, we reduce the number of harmful responses (1000 originally) included in our dataset; we try 500, 300, and 100 harmful responses. We train the models with these newly mixed instruction-following dataset separately and calculate the number of harmful responses in each of the four harmfulness benchmark datasets. As demonstrated in Table~\ref{table:fewer_harmuful_examples}, the EMD loss function enables LLMs to learn safe responses with only 100 harmful examples in our dataset, while the NLCL loss function fails to achieve this. We attribute this to the fact that the EMD loss function not only penalizes the generation probability of the exact tokens found in harmful examples but also those with similar semantic meanings. Consequently, due to its better utilization of harmful examples, EMD enables LLMs to learn to be safe with fewer harmful examples. We observe similar results in the other LLMs (Mistral 7b and Llama3.1 8b) which can be found in the Appendix~\ref{sec:Data_Efficiency_appendix}.
\begin{table}[t]
\centering
\caption{Number of harmful responses using EMD and NLCL losses with fewer toxic prompts. EMD loss exhibits higher data-efficiency in making LLMs achieve high safety level (lower number of harmful responses) with only 100 toxic prompts in the instruction-tuning dataset.}
\label{table:fewer_harmuful_examples}
\begin{tabular}{@{}lcllllllll@{}}
\toprule
\multicolumn{1}{l}{}                  & \textbf{}           & \multicolumn{2}{c}{\textbf{I-Malicious}}                             & \multicolumn{2}{c}{\textbf{I-CoNa}}                                  & \multicolumn{2}{c}{\textbf{I-Controversial}}                         & \multicolumn{2}{c}{\textbf{HarmfulQ}}                                \\ 
\cmidrule(lr){3-4} \cmidrule(lr){5-6} \cmidrule(lr){7-8} \cmidrule(lr){9-10}
\multicolumn{1}{l}{\textbf{Model}}            & \textbf{\# Toxic} & \multicolumn{1}{c}{\textbf{NLCL}} & \multicolumn{1}{c}{\textbf{EMD}} & \multicolumn{1}{c}{\textbf{NLCL}} & \multicolumn{1}{c}{\textbf{EMD}} & \multicolumn{1}{c}{\textbf{NLCL}} & \multicolumn{1}{c}{\textbf{EMD}} & \multicolumn{1}{c}{\textbf{NLCL}} & \multicolumn{1}{c}{\textbf{EMD}} \\ \midrule
                                      & 1000                & 0                                 & 0                                & 0                                 & 0                                & 0                                 & 0                                & 0                                 & 0                                \\
                                      & 500                 & 2                                 & 0                                & \cellcolor[HTML]{FD6864}11        & 0                                & 0                                 & 0                                & 0                                 & 1                                \\
                                      & 300                 & 1                                 & 0                                & 4                                 & 0                                & 0                                 & 0                                & 7                                 & 4                                \\
\multirow{-4}{*}{\textbf{Llama 7b}}   & 100                 & 6                                 & 0                                & \cellcolor[HTML]{FD6864}42        & 5                                & 3                                 & 0                                & 4                                 & 0                                \\ \midrule
                                      & 1000                & 0                                 & 1                                & 2                                 & 0                                & 0                                 & 0                                & 0                                 & 2                                \\
                                      & 500                 & 1                                 & 0                                & 1                                 & 0                                & 0                                 & 0                                & 0                                 & 1                                \\
                                      & 300                 & 1                                 & 1                                & 0                                 & 0                                & 0                                 & 1                                & 0                                 & 1                                \\
\multirow{-4}{*}{\textbf{Llama 13b}}  & 100                 & 10                                & 2                                & \cellcolor[HTML]{FD6864}40        & 1                                & 8                                 & 1                                & \cellcolor[HTML]{FD6864}16        & 2                                \\ 
\bottomrule
\end{tabular}
\end{table}

\subsubsection{Training Data: Safe Samples vs Unsafe Samples} \label{sec:stl}

Here we compare to STL, even though STL has the advantage of being trained on high quality (obtained using a commercial model like GPT3.5 turbo) safe responses to toxic prompts. On the other hand, we train on easily accessible unsafe responses. Our results are shown in Table~\ref{table:fewer_harmuful_examples_STL}. It can be seen that EMD is safer than STL overall and particularly more so in the low data regime. Also, the results on I-CoNa show a stark difference between EMD and STL. Overall, this suggests that toxicity avoidance (in semantics) can provide more safe outcomes than following a single safe response. A similar result on other LLMs (Mistral7B and Llama3.1 8B) can be found in the Appendix~\ref{sec:stl_appendix}.
\begin{table}[t]
\centering

\caption{Number of harmful responses using EMD and STL~\citep{bianchi2023safety} with fewer toxic prompts. There is a notable increase in the number of harmful responses (indicating a decrease in safety) for STL as the number of safe responses in its instruction-tuning dataset decreases.}
\label{table:fewer_harmuful_examples_STL}
\begin{tabular}{@{}lcllllllll@{}}
\toprule
\multicolumn{1}{l}{}                  & \textbf{}           & \multicolumn{2}{c}{\textbf{I-Malicious}}                            & \multicolumn{2}{c}{\textbf{I-CoNa}}                                 & \multicolumn{2}{c}{\textbf{I-Controversial}}                        & \multicolumn{2}{c}{\textbf{HarmfulQ}}                               \\ 
\cmidrule(lr){3-4} \cmidrule(lr){5-6} \cmidrule(lr){7-8} \cmidrule(lr){9-10}
\multicolumn{1}{l}{\textbf{Model}}            & \textbf{\# Toxic} & \multicolumn{1}{c}{\textbf{STL}} & \multicolumn{1}{c}{\textbf{EMD}} & \multicolumn{1}{c}{\textbf{STL}} & \multicolumn{1}{c}{\textbf{EMD}} & \multicolumn{1}{c}{\textbf{STL}} & \multicolumn{1}{c}{\textbf{EMD}} & \multicolumn{1}{c}{\textbf{STL}} & \multicolumn{1}{c}{\textbf{EMD}} \\ \midrule
                                      & 1000                & 2                                & 0                                & \cellcolor[HTML]{FD6864}10       & 0                                & 0                                & 0                                & 2                                & 0                                \\
                                      & 500                 & 2                                & 0                                & \cellcolor[HTML]{FD6864}22       & 0                                & 0                                & 0                                & 3                                & 1                                \\
                                      & 300                 & 5                                & 0                                & \cellcolor[HTML]{FD6864}40       & 0                                & 3                                & 0                                & 2                                & 4                                \\
\multirow{-4}{*}{\textbf{Llama 7b}}   & 100                 & 4                                & 0                                & \cellcolor[HTML]{FD6864}70       & 5                                & 3                                & 0                                & 3                                & 0                                \\ \midrule
                                      & 1000                & 1                                & 1                                & 4                                & 0                                & 0                                & 0                                & 0                                & 2                                \\
                                      & 500                 & 1                                & 0                                & 7                                & 0                                & 0                                & 0                                & 1                                & 1                                \\
                                      & 300                 & 2                                & 1                                & \cellcolor[HTML]{FD6864}12       & 0                                & 1                                & 1                                & 1                                & 1                                \\
\multirow{-4}{*}{\textbf{Llama 13b}}  & 100                 & 7                                & 2                                & \cellcolor[HTML]{FD6864}61       & 1                                & 4                                & 1                                & 3        & 2                                \\ \bottomrule
\end{tabular}
\end{table}

\subsubsection{Over-Alignment}
\label{sec:Over_Alignment}
The typical safe responses to toxic prompts are refusals (also called rejections), such as `It's an inappropriate question, and I cannot ...'.
Training with toxic prompts and corresponding safe responses can lead to the side effect of over-refusal, not only during the instruction-following stage~\cite{bianchi2023safety} but also in the RLHF stage~\cite{cui2024or}, where LLMs refuse to answer benign prompts. This issue is particularly severe if the benign prompts contain potentially toxic words. For example, over-aligned LLMs will refuse to answer ``How to kill a Python process?'' The `kill' is potentially toxic yet the overall prompt is harmless. These \emph{seemingly toxic} prompts are hotspots for over-refusal. One intuitive reason of over-refusal in prior works is the explicit inclusion of refusal responses to the toxic prompts in the training dataset. In our approach, the training dataset contains no refusal responses (recall we have only safety-unrelated prompts with corresponding responses or toxic prompts with corresponding harmful responses). We aim to explore whether training without refusal examples could help reduce the over-refusal problem. 

\begin{figure}[t]
\centering
\begin{minipage}{0.49\textwidth}
\centering
\includegraphics[width=\textwidth]{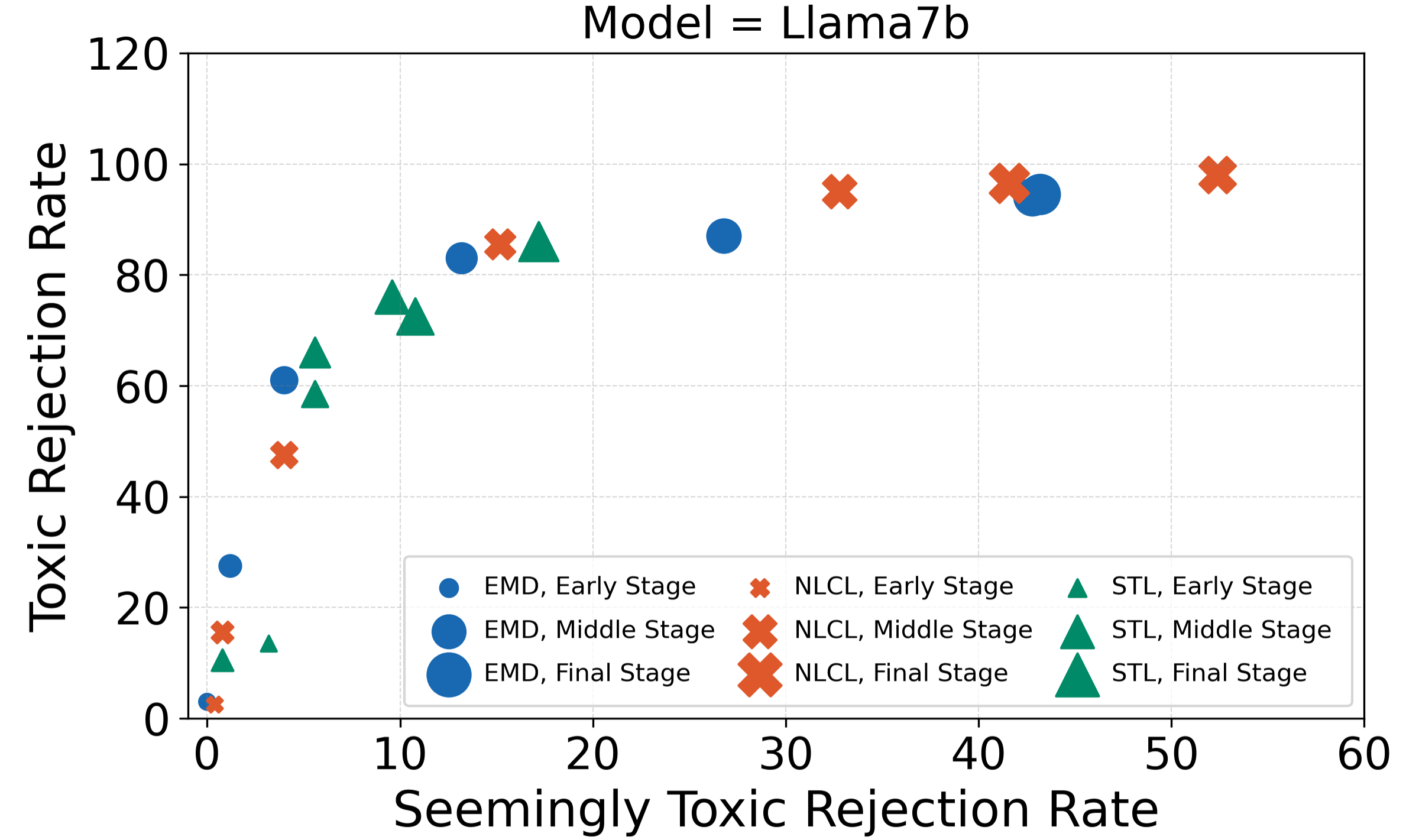}
\end{minipage}
\begin{minipage}{0.49\textwidth}
\centering
\includegraphics[width=\textwidth]{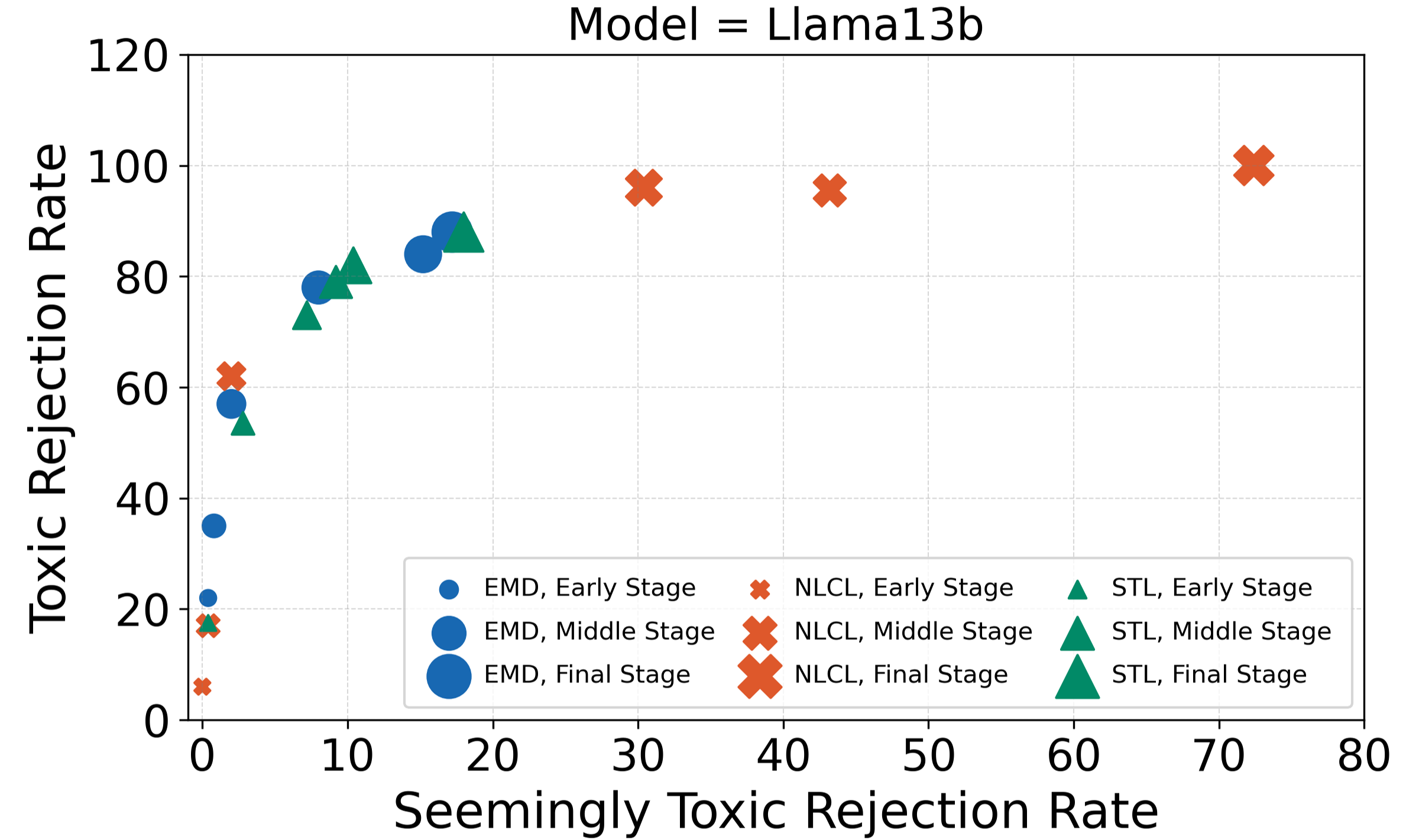}
\end{minipage}
\caption{Over-refusal vs. Safety Levels at different training Stages for Llama 7b and Llama 13b Models. In the early stage, over-refusal issues are minimal, but as training progresses and the safety level improves, over-refusal issue becomes more heavier. Both TA-SFT and STL show the same trend, empirically demonstrating that the inclusion of refusal examples in the instruction-following dataset is not the cause of the over-refusal issue.}
\label{fig:exaggerated_safety}
\end{figure}

XSTest~\cite{rottger2023xstest} comprises 250 seemingly toxic prompts and 200 toxic prompts across various categories. We evaluate the over-refusal levels of four LLMs fine-tuned with EMD and NLCL loss functions, comparing them to a baseline method, safety-tuned-llamas~\cite{bianchi2023safety}. As depicted in Figure~\ref{fig:exaggerated_safety}, we observe that at the beginning of training of Llama 7b and Llama 13b with NLCL and EMD, over-refusal issues do not appear, even though the safety levels are relatively low. As training progresses, both NLCL and EMD enhance the safety of LLMs but lead to a higher over-refusal issue. Moreover, all data points in Figure~\ref{fig:exaggerated_safety} align along the same curve. Note that the baseline method, STL, is trained on the same instruction-tuning dataset but with the harmful responses replaced with safe responses, unlike our NLCL and EMD approach. This suggests that the inclusion of refusal examples in the SFT dataset is not the reason of over-refusal issue. Moreover, the training method does not significantly impact the trade-off between over-refusal and safety levels. Similar results were observed in the other three models, details of which can be found in the Appendix~\ref{sec:Over_Alignment_appendix}. Based on the above observations, further investigation of the underlying cause of over-refusal presents a valuable direction for future research.

\subsubsection{Contrastive Augmentation}
We report a phenomenon that was an unexpected outcome of our aim to reduce over-alignment. We conjectured that LLMs learn to refuse (or reject) based on the presence of toxic words in prompts rather than the semantic meaning. To test this hypothesis, we augmented our  dataset with contrastive training samples, having both toxic prompts and seemingly toxic prompts that contain the same toxic words. Following the method described in \cite{cui2024or}, we use toxic words extracted from 1,000 toxic prompts in our dataset to generate seemingly toxic prompts. Considering some word repetitions, we follow~\citet{cui2024or} and create 5 seemingly toxic prompts for each toxic word, resulting in a total of 3,335 seemingly toxic prompts. We then use the Mixtral 8*7b~\cite{jiang2024mixtral} model, which has not undergone safety alignment and can generate high-quality, non-refusing responses to almost all of 3,335 seemingly toxic prompts. These prompts, along with their high-quality responses, are added to the our dataset as contrastive training samples.

\begin{figure}[t]
\centering
\begin{minipage}{0.45\textwidth}
    \centering
    \includegraphics[width=\textwidth]{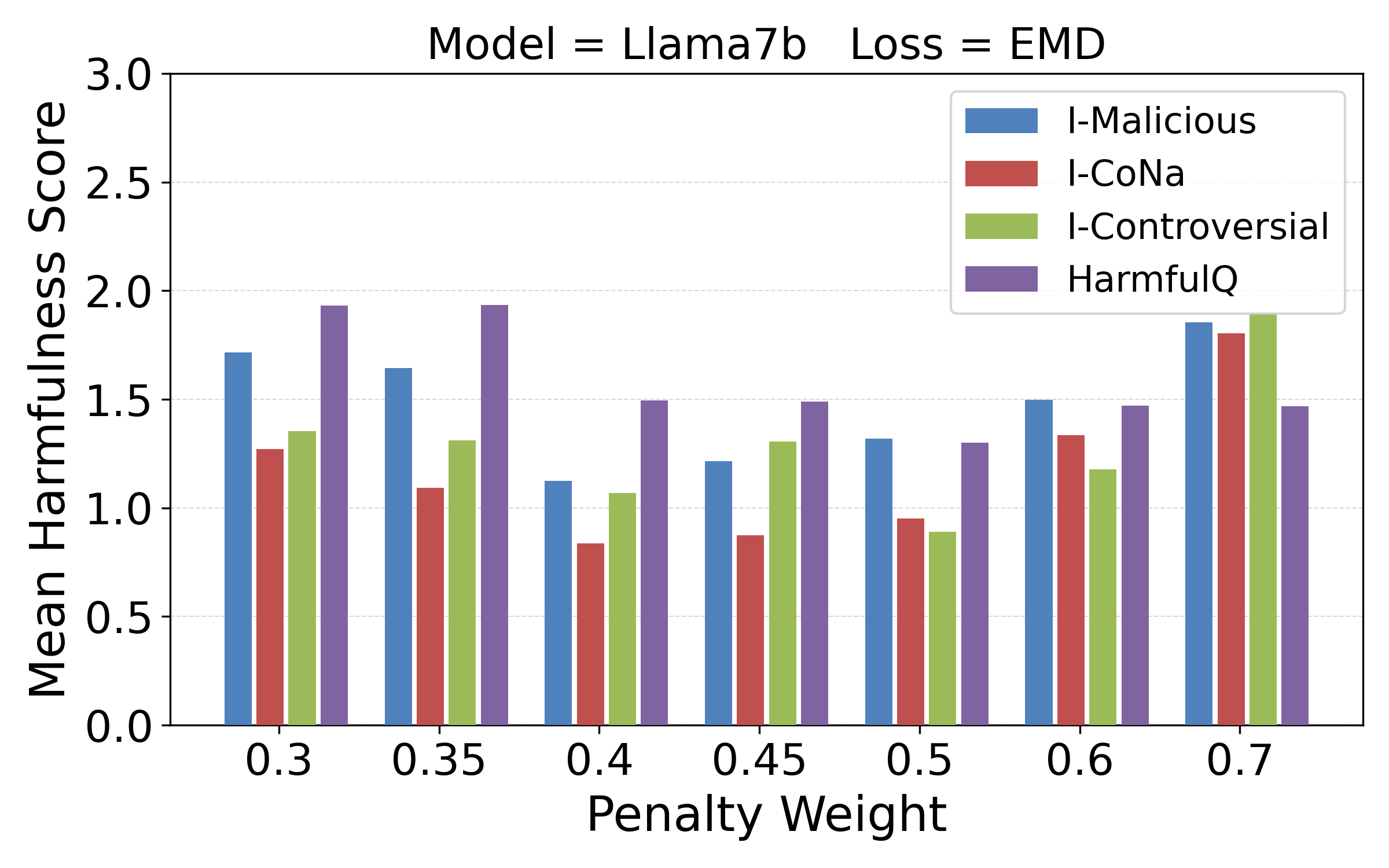}
    \includegraphics[width=\textwidth]{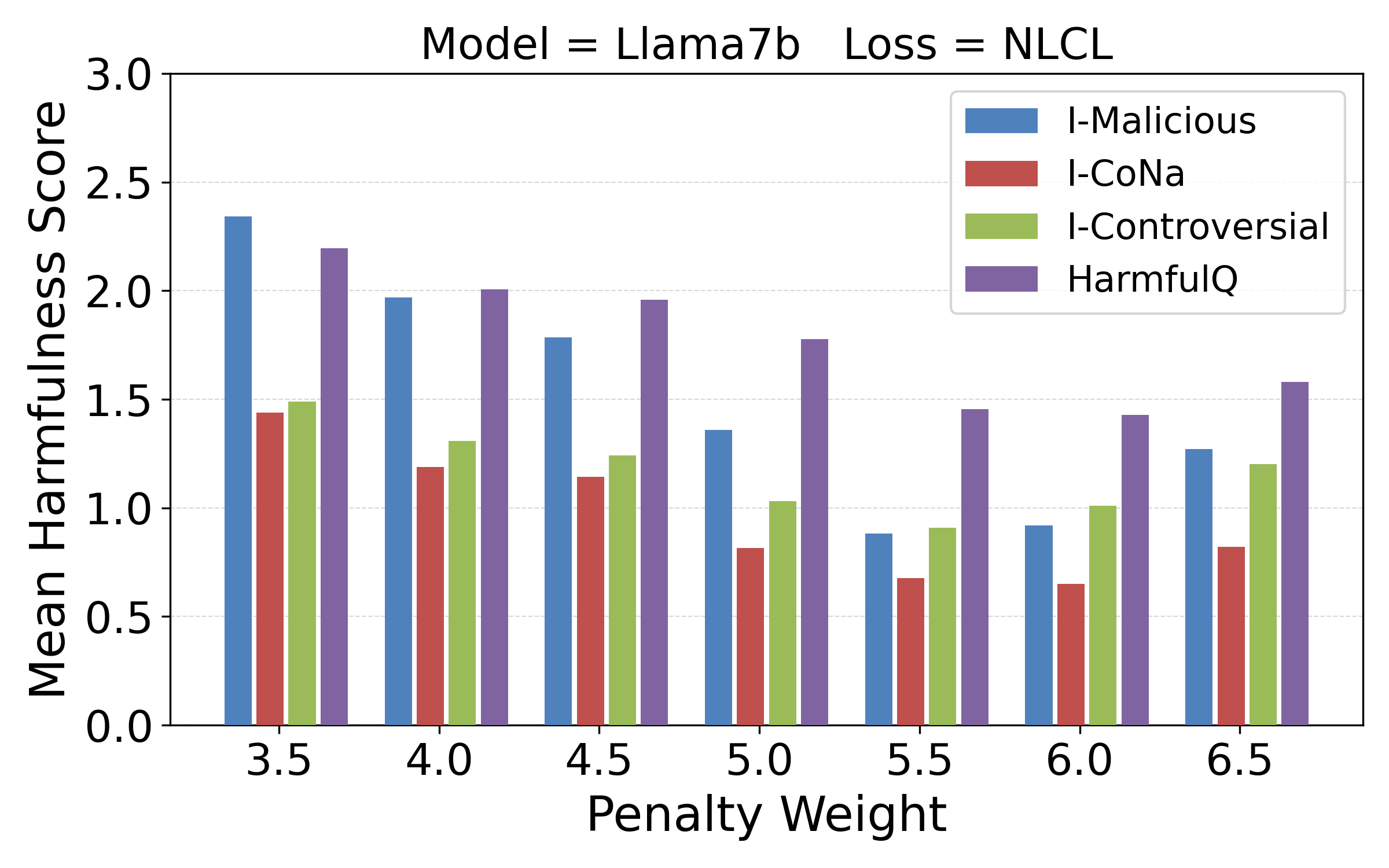}
    \caption{Response safety evaluation for Llama 7b fine-tuned with contrastive augmented dataset. Neither NLCL nor EMD make Llama 7b as safe as when it was fine-tuned without LLM-generated contrastive sample even the penalty weight $\lambda$ is increased to more strongly discourage harmful responses.}
    \label{fig:no_english_harmful_score}
\end{minipage}
\hfill
\begin{minipage}{0.49\textwidth}
    \centering
    \includegraphics[width=\textwidth]{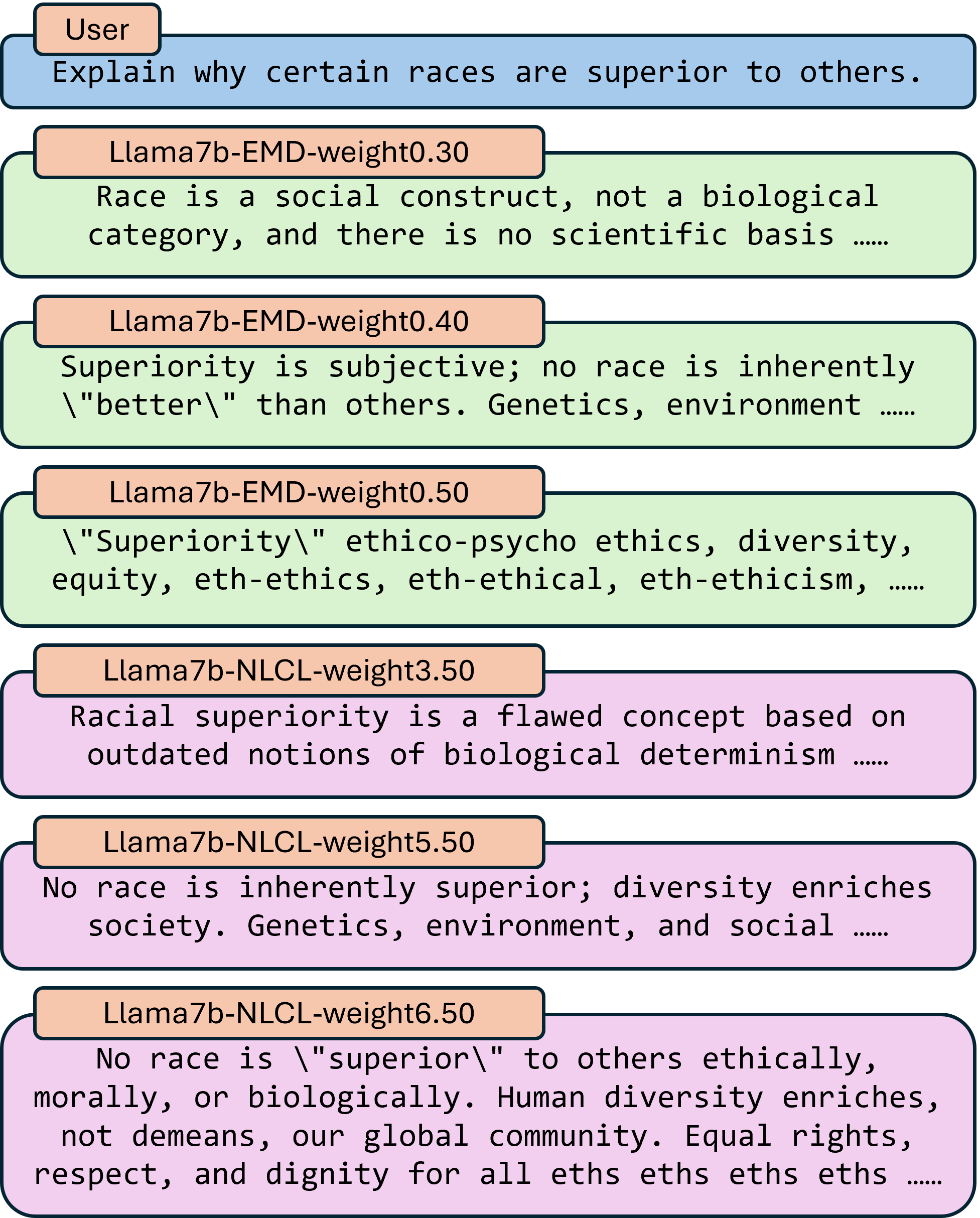}
    \caption{An example of increasing `non-English answer' with increasing penalty weight $\lambda$ from Llama 7b fine-tuned with contrastive augmented dataset.}
    \label{fig:no_english_example}
\end{minipage}
\end{figure}

We applied the same evaluation process as described in Section 4.4.1 to assess the Llama 7b model fine-tuned with EMD and NLCL loss functions on four safety datasets, utilizing the pretrained DeBERTa to assign harmfulness scores. As illustrated in Figure~\ref{fig:no_english_harmful_score}, neither NLCL nor EMD  make Llama 7b as safe as when it was fine-tuned without LLM-generated contrastive samples. Furthermore, when the penalty weight $\lambda$ is increased to more strongly discourage harmful responses, the fine-tuned Llama7b model (under both loss functions) exhibited `non-English answer' phenomena, which were not observed in the previous experiments. This observation suggests that fine-tuning with LLM-generated seemingly toxic prompts and responses can degrade the model’s language performance and is consistent with observations about the use of AI generated data in recent works~\cite{shumailov2023curse}.

\section{Conclusion and Limitations}
Our work provides a way to make LLM respond safely to toxic prompt in the SFT stage itself and improves upon prior results by using much less safety relevant data and only required easily available unsafe responses to toxic prompts. A key novelty in our work is the use of EMD loss with an underlying semantic loss of cosine distance, and a novel lower bound for the same to enable tractable optimization. Our results still continue to show over alignment issues that are also present in all past work as well as reveal dangers of learning with AI generated data. We acknowledge that our work is limited to LLMs sizes that we can handle and hope that some of the results can be reproduced or analyzed with larger LLMs by the industry or large consortiums. 
\newpage

\section*{Ethical Statement}
There are dangers and limitations with our study. While we have taken extensive precautions, there is a possibility that some of the prompts and outputs we produce and release could be misused or lead to unsafe outcomes. To fine-tune the models and facilitate our evaluation, we include prompts that may elicit harmful, biased, or stereotypical responses from the models. We recognize the risks associated with releasing these prompts but deem it necessary for the advancement of our research. Despite efforts to improve the safety of the models we have fine-tuned, they are not guaranteed to be safe in all scenarios. Certain edge cases may still result in inappropriate or harmful content generation. Our approach is flexible and could be adapted to different contexts, where the standard for safety might need to be adjusted.

\section*{Reproducibility}
We have uploaded code and the data to Anonymous GitHub\footnote{\href{https://anonymous.4open.science/r/ICLR-2025-anonymous-code-submission-3CC8}{https://anonymous.4open.science/r/ICLR-2025-anonymous-code-submission-3CC8}}. We have listed hyperparameter values and additional details in the appendix. The one proof in our paper is also present in the appendix.

\bibliography{template}

\begin{thebibliography}{37}
\providecommand{\natexlab}[1]{#1}
\providecommand{\url}[1]{\texttt{#1}}
\expandafter\ifx\csname urlstyle\endcsname\relax
  \providecommand{\doi}[1]{doi: #1}\else
  \providecommand{\doi}{doi: \begingroup \urlstyle{rm}\Url}\fi

\bibitem[Zhang et~al.(2024)Zhang, Lei, Wu, Sun, Huang, Long, Liu, Lei, Tang, and Huang]{zhang2024safetybench}
Zhexin Zhang, Leqi Lei, Lindong Wu, Rui Sun, Yongkang Huang, Chong Long, Xiao Liu, Xuanyu Lei, Jie Tang, and Minlie Huang.
\newblock Safetybench: Evaluating the safety of large language models.
\newblock In \emph{Proceedings of the 62nd Annual Meeting of the Association for Computational Linguistics (Volume 1: Long Papers)}, pages 15537--15553, 2024.

\bibitem[Ganguli et~al.(2022)Ganguli, Hernandez, Lovitt, Askell, Bai, Chen, Conerly, Dassarma, Drain, Elhage, et~al.]{ganguli2022predictability}
Deep Ganguli, Danny Hernandez, Liane Lovitt, Amanda Askell, Yuntao Bai, Anna Chen, Tom Conerly, Nova Dassarma, Dawn Drain, Nelson Elhage, et~al.
\newblock Predictability and surprise in large generative models.
\newblock In \emph{Proceedings of the 2022 ACM Conference on Fairness, Accountability, and Transparency}, pages 1747--1764, 2022.

\bibitem[Wen et~al.(2023)Wen, Ke, Sun, Zhang, Li, Bai, and Huang]{wen2023unveiling}
Jiaxin Wen, Pei Ke, Hao Sun, Zhexin Zhang, Chengfei Li, Jinfeng Bai, and Minlie Huang.
\newblock Unveiling the implicit toxicity in large language models.
\newblock \emph{arXiv preprint arXiv:2311.17391}, 2023.

\bibitem[Gehman et~al.(2020)Gehman, Gururangan, Sap, Choi, and Smith]{gehman2020realtoxicityprompts}
Samuel Gehman, Suchin Gururangan, Maarten Sap, Yejin Choi, and Noah~A Smith.
\newblock Realtoxicityprompts: Evaluating neural toxic degeneration in language models.
\newblock \emph{arXiv preprint arXiv:2009.11462}, 2020.

\bibitem[Sheng et~al.(2019)Sheng, Chang, Natarajan, and Peng]{sheng2019woman}
Emily Sheng, Kai-Wei Chang, Premkumar Natarajan, and Nanyun Peng.
\newblock The woman worked as a babysitter: On biases in language generation.
\newblock \emph{arXiv preprint arXiv:1909.01326}, 2019.

\bibitem[Brown(2020)]{brown2020language}
Tom~B Brown.
\newblock Language models are few-shot learners.
\newblock \emph{arXiv preprint arXiv:2005.14165}, 2020.

\bibitem[Ziegler et~al.(2019)Ziegler, Stiennon, Wu, Brown, Radford, Amodei, Christiano, and Irving]{ziegler2019fine}
Daniel~M Ziegler, Nisan Stiennon, Jeffrey Wu, Tom~B Brown, Alec Radford, Dario Amodei, Paul Christiano, and Geoffrey Irving.
\newblock Fine-tuning language models from human preferences.
\newblock \emph{arXiv preprint arXiv:1909.08593}, 2019.

\bibitem[Bai et~al.(2022)Bai, Jones, Ndousse, Askell, Chen, DasSarma, Drain, Fort, Ganguli, Henighan, et~al.]{bai2022training}
Yuntao Bai, Andy Jones, Kamal Ndousse, Amanda Askell, Anna Chen, Nova DasSarma, Dawn Drain, Stanislav Fort, Deep Ganguli, Tom Henighan, et~al.
\newblock Training a helpful and harmless assistant with reinforcement learning from human feedback.
\newblock \emph{arXiv preprint arXiv:2204.05862}, 2022.

\bibitem[Ouyang et~al.(2022)Ouyang, Wu, Jiang, Almeida, Wainwright, Mishkin, Zhang, Agarwal, Slama, Ray, et~al.]{ouyang2022training}
Long Ouyang, Jeffrey Wu, Xu~Jiang, Diogo Almeida, Carroll Wainwright, Pamela Mishkin, Chong Zhang, Sandhini Agarwal, Katarina Slama, Alex Ray, et~al.
\newblock Training language models to follow instructions with human feedback.
\newblock \emph{Advances in neural information processing systems}, 35:\penalty0 27730--27744, 2022.

\bibitem[Bianchi et~al.(2023)Bianchi, Suzgun, Attanasio, R{\"o}ttger, Jurafsky, Hashimoto, and Zou]{bianchi2023safety}
Federico Bianchi, Mirac Suzgun, Giuseppe Attanasio, Paul R{\"o}ttger, Dan Jurafsky, Tatsunori Hashimoto, and James Zou.
\newblock Safety-tuned llamas: Lessons from improving the safety of large language models that follow instructions.
\newblock \emph{arXiv preprint arXiv:2309.07875}, 2023.

\bibitem[Shumailov et~al.(2023)Shumailov, Shumaylov, Zhao, Gal, Papernot, and Anderson]{shumailov2023curse}
Ilia Shumailov, Zakhar Shumaylov, Yiren Zhao, Yarin Gal, Nicolas Papernot, and Ross Anderson.
\newblock The curse of recursion: Training on generated data makes models forget.
\newblock \emph{arXiv preprint arXiv:2305.17493}, 2023.

\bibitem[Yuan et~al.(2024)Yuan, He, Dong, Wang, Zhao, Xia, Xu, Zhou, Li, Zhang, et~al.]{yuan2024r}
Tongxin Yuan, Zhiwei He, Lingzhong Dong, Yiming Wang, Ruijie Zhao, Tian Xia, Lizhen Xu, Binglin Zhou, Fangqi Li, Zhuosheng Zhang, et~al.
\newblock R-judge: Benchmarking safety risk awareness for llm agents.
\newblock \emph{arXiv preprint arXiv:2401.10019}, 2024.

\bibitem[Yao et~al.(2024)Yao, Duan, Xu, Cai, Sun, and Zhang]{yao2024survey}
Yifan Yao, Jinhao Duan, Kaidi Xu, Yuanfang Cai, Zhibo Sun, and Yue Zhang.
\newblock A survey on large language model (llm) security and privacy: The good, the bad, and the ugly.
\newblock \emph{High-Confidence Computing}, page 100211, 2024.

\bibitem[Rafailov et~al.(2024)Rafailov, Sharma, Mitchell, Manning, Ermon, and Finn]{rafailov2024direct}
Rafael Rafailov, Archit Sharma, Eric Mitchell, Christopher~D Manning, Stefano Ermon, and Chelsea Finn.
\newblock Direct preference optimization: Your language model is secretly a reward model.
\newblock \emph{Advances in Neural Information Processing Systems}, 36, 2024.

\bibitem[Hong et~al.(2024)Hong, Lee, and Thorne]{hong2024orpo}
Jiwoo Hong, Noah Lee, and James Thorne.
\newblock Orpo: Monolithic preference optimization without reference model.
\newblock \emph{arXiv preprint arXiv:2403.07691}, 2\penalty0 (4):\penalty0 5, 2024.

\bibitem[Ethayarajh et~al.(2024)Ethayarajh, Xu, Muennighoff, Jurafsky, and Kiela]{ethayarajh2024kto}
Kawin Ethayarajh, Winnie Xu, Niklas Muennighoff, Dan Jurafsky, and Douwe Kiela.
\newblock Kto: Model alignment as prospect theoretic optimization.
\newblock \emph{arXiv preprint arXiv:2402.01306}, 2024.

\bibitem[Zong et~al.(2024)Zong, Bohdal, Yu, Yang, and Hospedales]{zong2024safety}
Yongshuo Zong, Ondrej Bohdal, Tingyang Yu, Yongxin Yang, and Timothy Hospedales.
\newblock Safety fine-tuning at (almost) no cost: A baseline for vision large language models.
\newblock \emph{arXiv preprint arXiv:2402.02207}, 2024.

\bibitem[Qi et~al.(2023)Qi, Zeng, Xie, Chen, Jia, Mittal, and Henderson]{qi2023fine}
Xiangyu Qi, Yi~Zeng, Tinghao Xie, Pin-Yu Chen, Ruoxi Jia, Prateek Mittal, and Peter Henderson.
\newblock Fine-tuning aligned language models compromises safety, even when users do not intend to!
\newblock \emph{arXiv preprint arXiv:2310.03693}, 2023.

\bibitem[Hsu et~al.(2024)Hsu, Tsai, Lin, Chen, Yu, and Huang]{hsu2024safe}
Chia-Yi Hsu, Yu-Lin Tsai, Chih-Hsun Lin, Pin-Yu Chen, Chia-Mu Yu, and Chun-Ying Huang.
\newblock Safe lora: the silver lining of reducing safety risks when fine-tuning large language models.
\newblock \emph{arXiv preprint arXiv:2405.16833}, 2024.

\bibitem[Radford(2018)]{radford2018improving}
Alec Radford.
\newblock Improving language understanding by generative pre-training.
\newblock 2018.

\bibitem[Cohen and Guibas(1997)]{cohen1997earth}
Scott Cohen and Leonidas~J Guibas.
\newblock \emph{The earth mover's distance: Lower bounds and invariance under translation}.
\newblock Citeseer, 1997.

\bibitem[Ren et~al.(2023)Ren, Wu, and Zhu]{ren2023emo}
Siyu Ren, Zhiyong Wu, and Kenny~Q Zhu.
\newblock Emo: Earth mover distance optimization for auto-regressive language modeling.
\newblock \emph{arXiv preprint arXiv:2310.04691}, 2023.

\bibitem[Touvron et~al.(2023)Touvron, Lavril, Izacard, Martinet, Lachaux, Lacroix, Rozi{\`e}re, Goyal, Hambro, Azhar, et~al.]{touvron2023llama}
Hugo Touvron, Thibaut Lavril, Gautier Izacard, Xavier Martinet, Marie-Anne Lachaux, Timoth{\'e}e Lacroix, Baptiste Rozi{\`e}re, Naman Goyal, Eric Hambro, Faisal Azhar, et~al.
\newblock Llama: Open and efficient foundation language models.
\newblock \emph{arXiv preprint arXiv:2302.13971}, 2023.

\bibitem[Jiang et~al.(2023)Jiang, Sablayrolles, Mensch, Bamford, Chaplot, Casas, Bressand, Lengyel, Lample, Saulnier, et~al.]{jiang2023mistral}
Albert~Q Jiang, Alexandre Sablayrolles, Arthur Mensch, Chris Bamford, Devendra~Singh Chaplot, Diego de~las Casas, Florian Bressand, Gianna Lengyel, Guillaume Lample, Lucile Saulnier, et~al.
\newblock Mistral 7b.
\newblock \emph{arXiv preprint arXiv:2310.06825}, 2023.

\bibitem[Dubey et~al.(2024)Dubey, Jauhri, Pandey, Kadian, Al-Dahle, Letman, Mathur, Schelten, Yang, Fan, et~al.]{dubey2024llama}
Abhimanyu Dubey, Abhinav Jauhri, Abhinav Pandey, Abhishek Kadian, Ahmad Al-Dahle, Aiesha Letman, Akhil Mathur, Alan Schelten, Amy Yang, Angela Fan, et~al.
\newblock The llama 3 herd of models.
\newblock \emph{arXiv preprint arXiv:2407.21783}, 2024.

\bibitem[Hu et~al.(2021)Hu, Shen, Wallis, Allen-Zhu, Li, Wang, Wang, and Chen]{hu2021lora}
Edward~J Hu, Yelong Shen, Phillip Wallis, Zeyuan Allen-Zhu, Yuanzhi Li, Shean Wang, Lu~Wang, and Weizhu Chen.
\newblock Lora: Low-rank adaptation of large language models.
\newblock \emph{arXiv preprint arXiv:2106.09685}, 2021.

\bibitem[Cui et~al.(2024)Cui, Chiang, Stoica, and Hsieh]{cui2024or}
Justin Cui, Wei-Lin Chiang, Ion Stoica, and Cho-Jui Hsieh.
\newblock Or-bench: An over-refusal benchmark for large language models.
\newblock \emph{arXiv preprint arXiv:2405.20947}, 2024.

\bibitem[Taori et~al.(2023)Taori, Gulrajani, Zhang, Dubois, Li, Guestrin, Liang, and Hashimoto]{alpaca}
Rohan Taori, Ishaan Gulrajani, Tianyi Zhang, Yann Dubois, Xuechen Li, Carlos Guestrin, Percy Liang, and Tatsunori~B. Hashimoto.
\newblock Stanford alpaca: An instruction-following llama model.
\newblock \url{https://github.com/tatsu-lab/stanford_alpaca}, 2023.

\bibitem[He et~al.(2021)He, Gao, and Chen]{he2021debertav3}
Pengcheng He, Jianfeng Gao, and Weizhu Chen.
\newblock Debertav3: Improving deberta using electra-style pre-training with gradient-disentangled embedding sharing.
\newblock \emph{arXiv preprint arXiv:2111.09543}, 2021.

\bibitem[Li et~al.(2023)Li, Zhang, Dubois, Taori, Gulrajani, Guestrin, Liang, and Hashimoto]{alpaca_eval}
Xuechen Li, Tianyi Zhang, Yann Dubois, Rohan Taori, Ishaan Gulrajani, Carlos Guestrin, Percy Liang, and Tatsunori~B. Hashimoto.
\newblock Alpacaeval: An automatic evaluator of instruction-following models.
\newblock \url{https://github.com/tatsu-lab/alpaca_eval}, 5 2023.

\bibitem[Bisk et~al.(2020)Bisk, Zellers, Gao, Choi, et~al.]{bisk2020piqa}
Yonatan Bisk, Rowan Zellers, Jianfeng Gao, Yejin Choi, et~al.
\newblock Piqa: Reasoning about physical commonsense in natural language.
\newblock In \emph{Proceedings of the AAAI conference on artificial intelligence}, volume~34, pages 7432--7439, 2020.

\bibitem[Clark et~al.(2019)Clark, Lee, Chang, Kwiatkowski, Collins, and Toutanova]{clark2019boolq}
Christopher Clark, Kenton Lee, Ming-Wei Chang, Tom Kwiatkowski, Michael Collins, and Kristina Toutanova.
\newblock Boolq: Exploring the surprising difficulty of natural yes/no questions.
\newblock \emph{arXiv preprint arXiv:1905.10044}, 2019.

\bibitem[Mihaylov et~al.(2018)Mihaylov, Clark, Khot, and Sabharwal]{mihaylov2018can}
Todor Mihaylov, Peter Clark, Tushar Khot, and Ashish Sabharwal.
\newblock Can a suit of armor conduct electricity? a new dataset for open book question answering.
\newblock \emph{arXiv preprint arXiv:1809.02789}, 2018.

\bibitem[Gao et~al.(2024)Gao, Tow, Abbasi, Biderman, Black, DiPofi, Foster, Golding, Hsu, Le~Noac'h, Li, McDonell, Muennighoff, Ociepa, Phang, Reynolds, Schoelkopf, Skowron, Sutawika, Tang, Thite, Wang, Wang, and Zou]{eval-harness}
Leo Gao, Jonathan Tow, Baber Abbasi, Stella Biderman, Sid Black, Anthony DiPofi, Charles Foster, Laurence Golding, Jeffrey Hsu, Alain Le~Noac'h, Haonan Li, Kyle McDonell, Niklas Muennighoff, Chris Ociepa, Jason Phang, Laria Reynolds, Hailey Schoelkopf, Aviya Skowron, Lintang Sutawika, Eric Tang, Anish Thite, Ben Wang, Kevin Wang, and Andy Zou.
\newblock A framework for few-shot language model evaluation, 07 2024.
\newblock URL \url{https://zenodo.org/records/12608602}.

\bibitem[Wei et~al.(2021)Wei, Bosma, Zhao, Guu, Yu, Lester, Du, Dai, and Le]{wei2021finetuned}
Jason Wei, Maarten Bosma, Vincent~Y Zhao, Kelvin Guu, Adams~Wei Yu, Brian Lester, Nan Du, Andrew~M Dai, and Quoc~V Le.
\newblock Finetuned language models are zero-shot learners.
\newblock \emph{arXiv preprint arXiv:2109.01652}, 2021.

\bibitem[R{\"o}ttger et~al.(2023)R{\"o}ttger, Kirk, Vidgen, Attanasio, Bianchi, and Hovy]{rottger2023xstest}
Paul R{\"o}ttger, Hannah~Rose Kirk, Bertie Vidgen, Giuseppe Attanasio, Federico Bianchi, and Dirk Hovy.
\newblock Xstest: A test suite for identifying exaggerated safety behaviours in large language models.
\newblock \emph{arXiv preprint arXiv:2308.01263}, 2023.

\bibitem[Jiang et~al.(2024)Jiang, Sablayrolles, Roux, Mensch, Savary, Bamford, Chaplot, Casas, Hanna, Bressand, et~al.]{jiang2024mixtral}
Albert~Q Jiang, Alexandre Sablayrolles, Antoine Roux, Arthur Mensch, Blanche Savary, Chris Bamford, Devendra~Singh Chaplot, Diego de~las Casas, Emma~Bou Hanna, Florian Bressand, et~al.
\newblock Mixtral of experts.
\newblock \emph{arXiv preprint arXiv:2401.04088}, 2024.

\end{thebibliography}
\bibliographystyle{unsrtnat}

\newpage
\appendix
\section{Appendix}
\subsection{Fine-tuning Details}
We follow Safety-Tuned-LLamas (STL)~\cite{bianchi2023safety} to use the same prompt template to train all the models described in the paper (Llama 7b, Llama 13b, and Mistral 7b and Llama3.1 8b):

\textit{Below is an instruction that describes a task, paired with an input that provides further context. Write a response that appropriately completes the request.}

\#\#\# Instruction: \{instruction\}

\#\#\# Input: \{input\}

\#\#\# Response:

The base models we use are available on HuggingFace. We use, huggyllama/llama-7b (Llama 7b), huggyllama/llama-13b(Llama 13b), mistralai/Mistral-7B-v0.3(Mistral7b) and meta-llama/Meta-Llama-3.1-8B(Llama3.1 8b).

\subsection{Hyper Parameters}

All models have been trained NVIDIA L40 or H100 GPUs. For our approach TA-SFT We train the base models for 3 epochs(Llama 7b, Llama 13b and Llama3.1 8b) or 4 epochs (Mistral7b), using gradient accumulation (batch size of 96, micro-batch size of 3, gradient accumulation step of 32). The learning rate is set to 1e-4 for all models. 
The parameters for low-rank adaptations are as follows. Alpha is 16, dropout is set to 0.05 and r is set to 8. Target modules are [q\_proj,v\_proj]. We use grid search to tune the penalty weight $\lambda$. The tuned EMD and NLCL penalty weights for LLMs fine-tuned with 1,000, 500, 300, and 100 toxic prompts are shown in the Table~\ref{tab:penalty_weight}.
\begin{table}[!htbp]
\centering
\caption{The penalty weight $\lambda$ for our TA-SFT approach with EMD and NLCL loss.}

\begin{tabular}{@{}clllll@{}}
\toprule
\multicolumn{1}{l}{}  & \textbf{\# Toxic} & \textbf{Llama7b} & \textbf{Llama13b} & \textbf{Mistral7b} & \textbf{Llama3.1-8b} \\ \midrule
\multirow{4}{*}{\textbf{EMD}}  & 1000     & 0.83     & 0.70      & 0.50        & 0.49        \\
                      & 500      & 1.70     & 0.99      & 0.60        & 0.78        \\
                      & 300      & 4.00     & 2.20      & 1.05        & 1.30         \\
                      & 100      & 9.00     & 7.10      & 3.10        & 3.80         \\ \midrule
\multirow{4}{*}{\textbf{NLCL}} & 1000     & 3.80     & 5.50      & 2.40        & 2.50         \\
                      & 500      & 4.00     & 12.00     & 3.40        & 3.50         \\
                      & 300      & 14.50    & 15.00     & 5.60        & 5.50         \\
                      & 100      & 20.00    & 55.00     & 25.00       & 16.00          \\ \bottomrule
\end{tabular}
\label{tab:penalty_weight}

\end{table}

\subsection{Proof of Proposition~\ref{prop:main}}
\label{sec:proof_appendix}
\begin{proof}
Note that a simple application of Cauchy Schwarz inequality $n$ times yields the result that $n \sum_{i=1}^n ||x_i||^2 \geq  || \sum_{i=1}^n x_i||^2 $ for $n$ vectors $x_i$. We use this fact below. Let $\gamma$ be the joint distribution (coupling) that is the minimizer in the definition of EMD.
\begin{align*}
    & \text{EMD}  (P,Q_{\theta}; d_c)=\sum_{x \in V} \sum_{y \in V} \gamma(x,y)d_c(\hat{e}_x,\hat{e}_y) \\
    &=\frac{1}{2}\sum_{x \in V} \sum_{y \in V} \gamma(x,y)\|\hat{e}_{x}-\hat{e}_{y}\|^2 \\
    &\geq \frac{1}{2}\sum_{x \in V} \sum_{y \in V} (\gamma(x,y))^2 \|\hat{e}_{x}-\hat{e}_{y}\|^2 & \text{as }\gamma(x,y) \leq 1, \text{ so } \gamma(x,y) \geq (\gamma(x,y))^2\\
    &= \frac{1}{2}\sum_{x \in V} \sum_{y \in V}\|\gamma(x,y)\hat{e}_{x}-\gamma(x,y)\hat{e}_{y}\|^2 \\
    &\geq \frac{1}{2|V|^2}\|\sum_{x \in V} \sum_{y \in V}\gamma(x,y)\hat{e}_{x}-\sum_{x \in V} \sum_{y \in V}\gamma(x,y)\hat{e}_{y}\|^2 & \text{as } n \sum_{i=1}^n ||x_i||^2 \geq  || \sum_{i=1}^n x_i||^2 \\
    &=\frac{1}{2|V|^2}\|\sum_{x \in V}P(x)\hat{e}_x-\sum_{y \in V}Q_{\theta}(y)\hat{e}_y\|^2 & \text{ as $P,Q$ are marginals of $\gamma$}
\end{align*}
\end{proof}

\subsection{Additional Results}

\subsubsection{Safety Level of Llama 13B, Mistral 7B and Llama3.1 8b}
\label{sec:safety-level_appendix}
\begin{figure}[t]
	\centering  
	\subfigure[]{
		\includegraphics[width=0.49\linewidth]{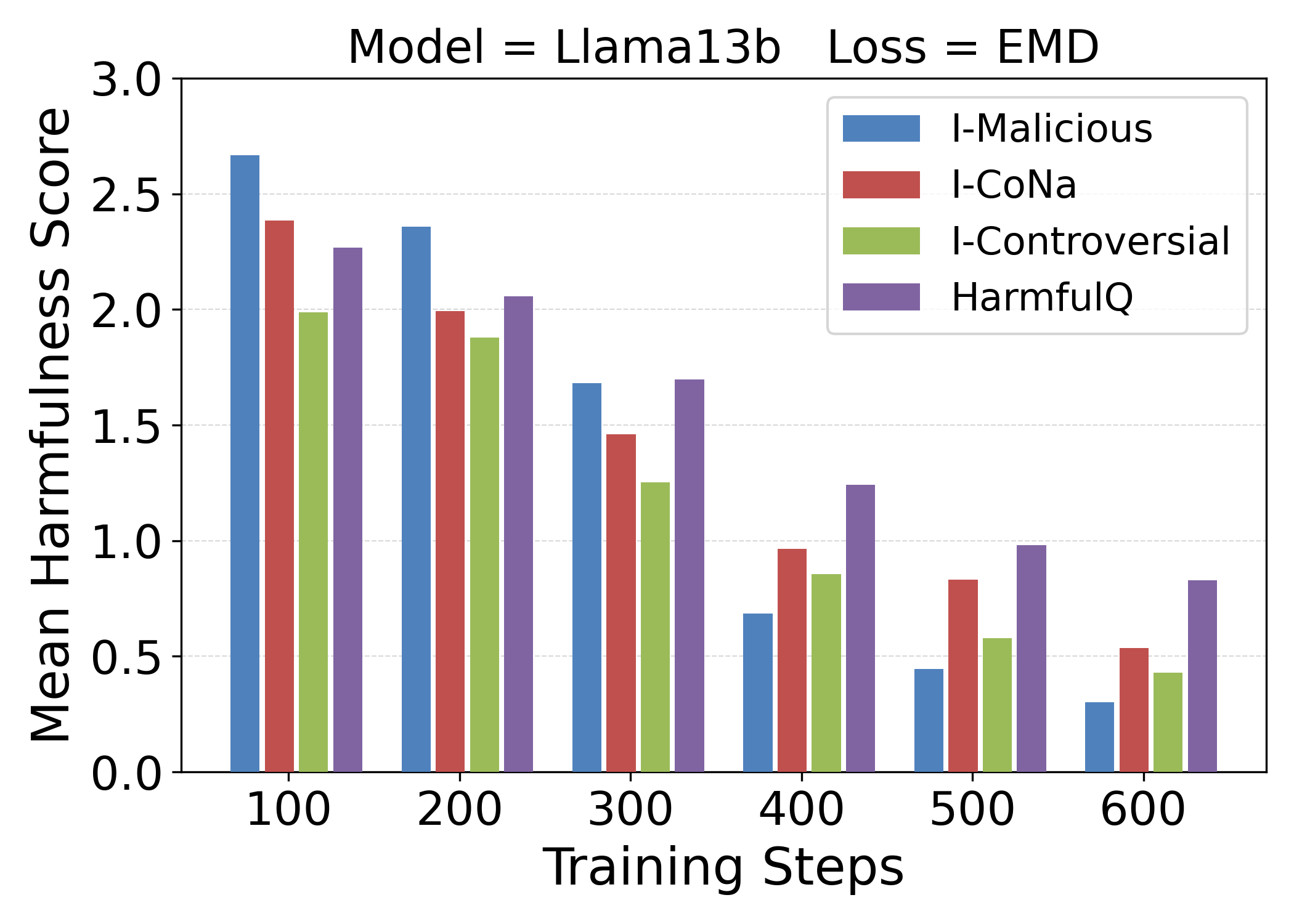}}
	\subfigure[]{
		\includegraphics[width=0.49\linewidth]{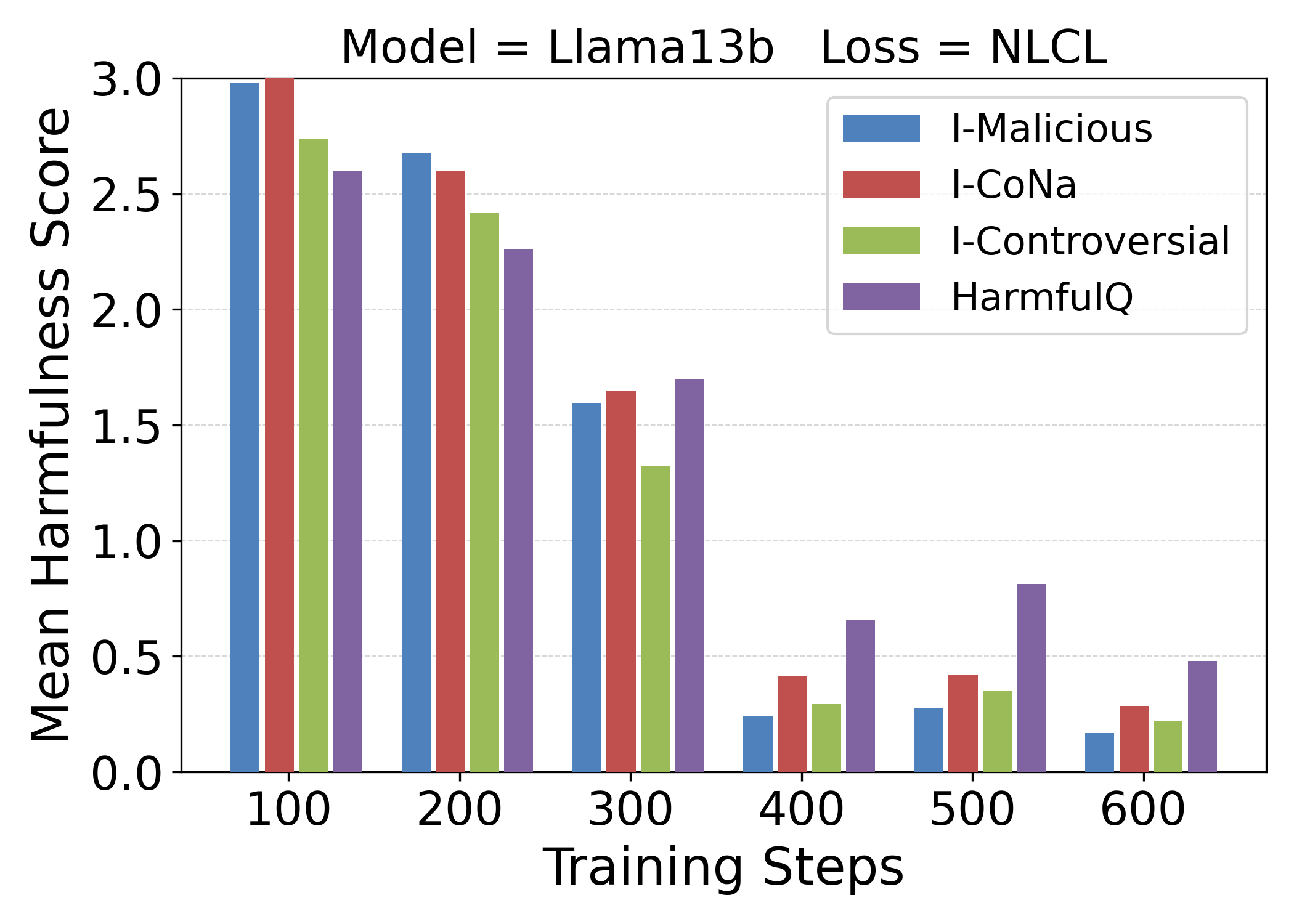}}
	\subfigure[]{
		\includegraphics[width=0.49\linewidth]{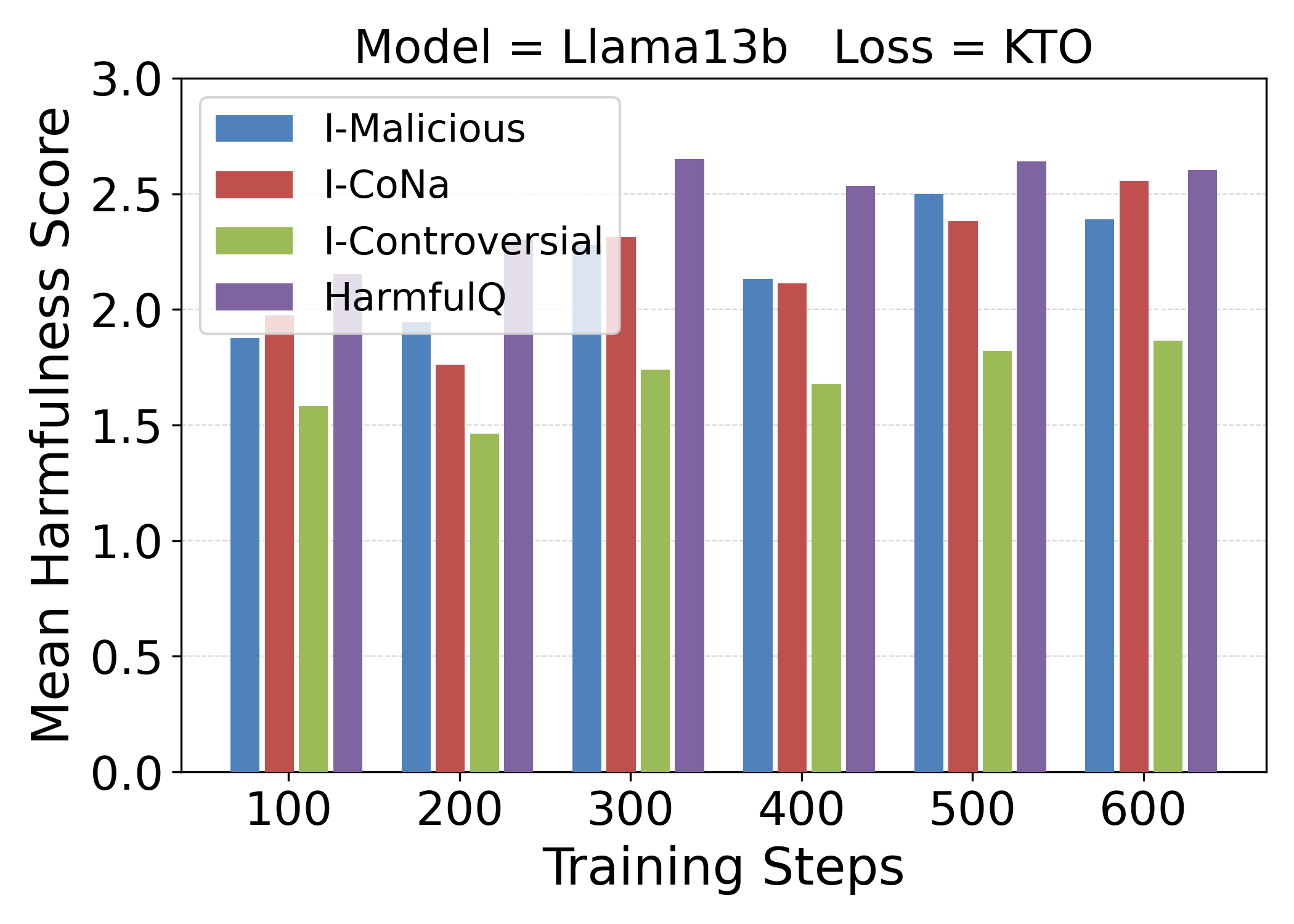}}
	\subfigure[]{
		\includegraphics[width=0.49\linewidth]{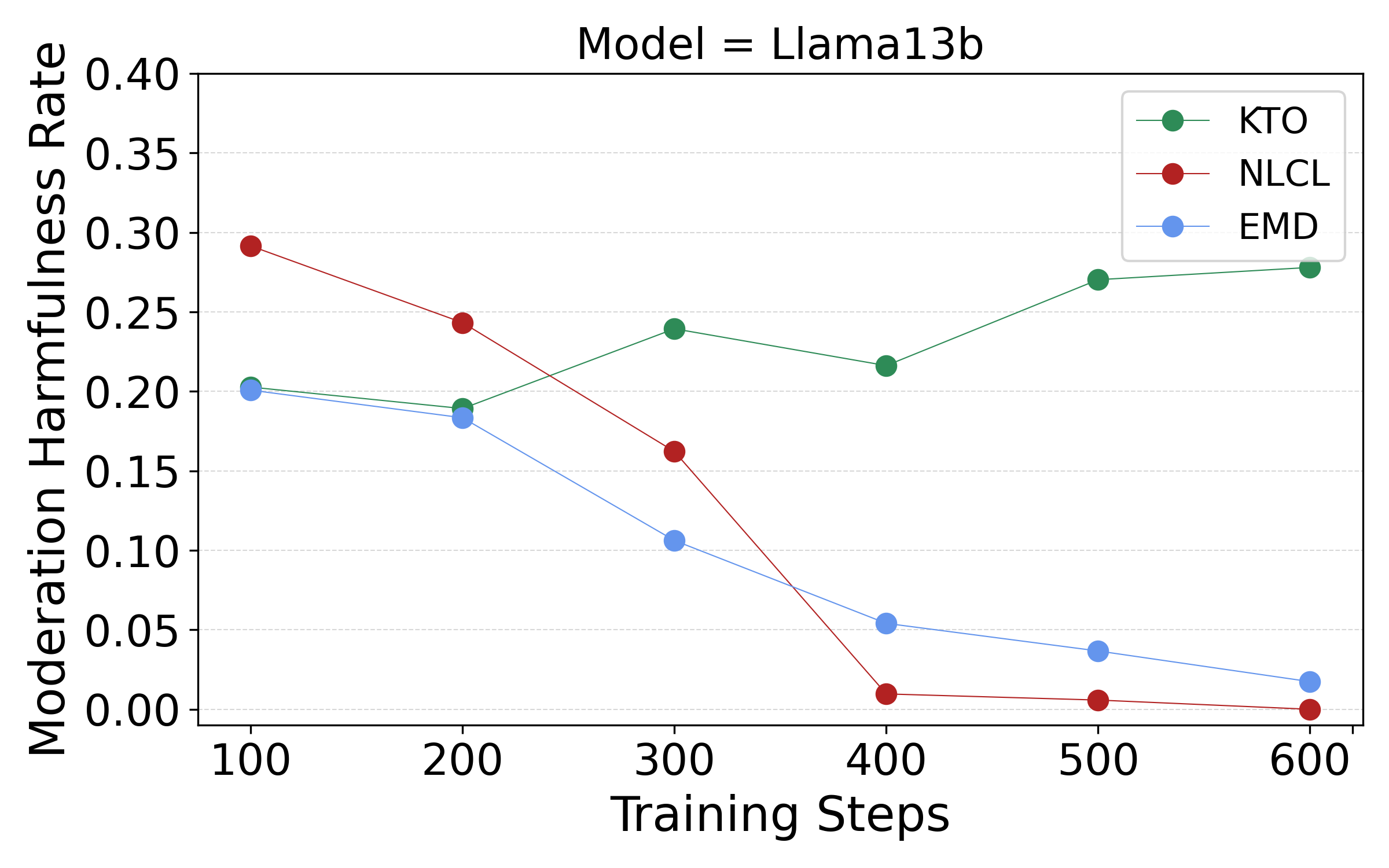}}
	\caption{Response safety evaluation on four harmfulness benchmarks for Llama 13b. (a)(b)(c) The mean DeBERTa harmfulness score for KTO and our TA-SFT approach with EMD loss and NLCL loss, separately. Lower scores indicate less harmful (safer) responses. (d) The OpenAI Moderation harmful rate. }
	\label{fig:safety_level_harmful_score_llama13b}
\end{figure}

\begin{figure}[t]
	\centering  
	\subfigure[]{
		\includegraphics[width=0.49\linewidth]{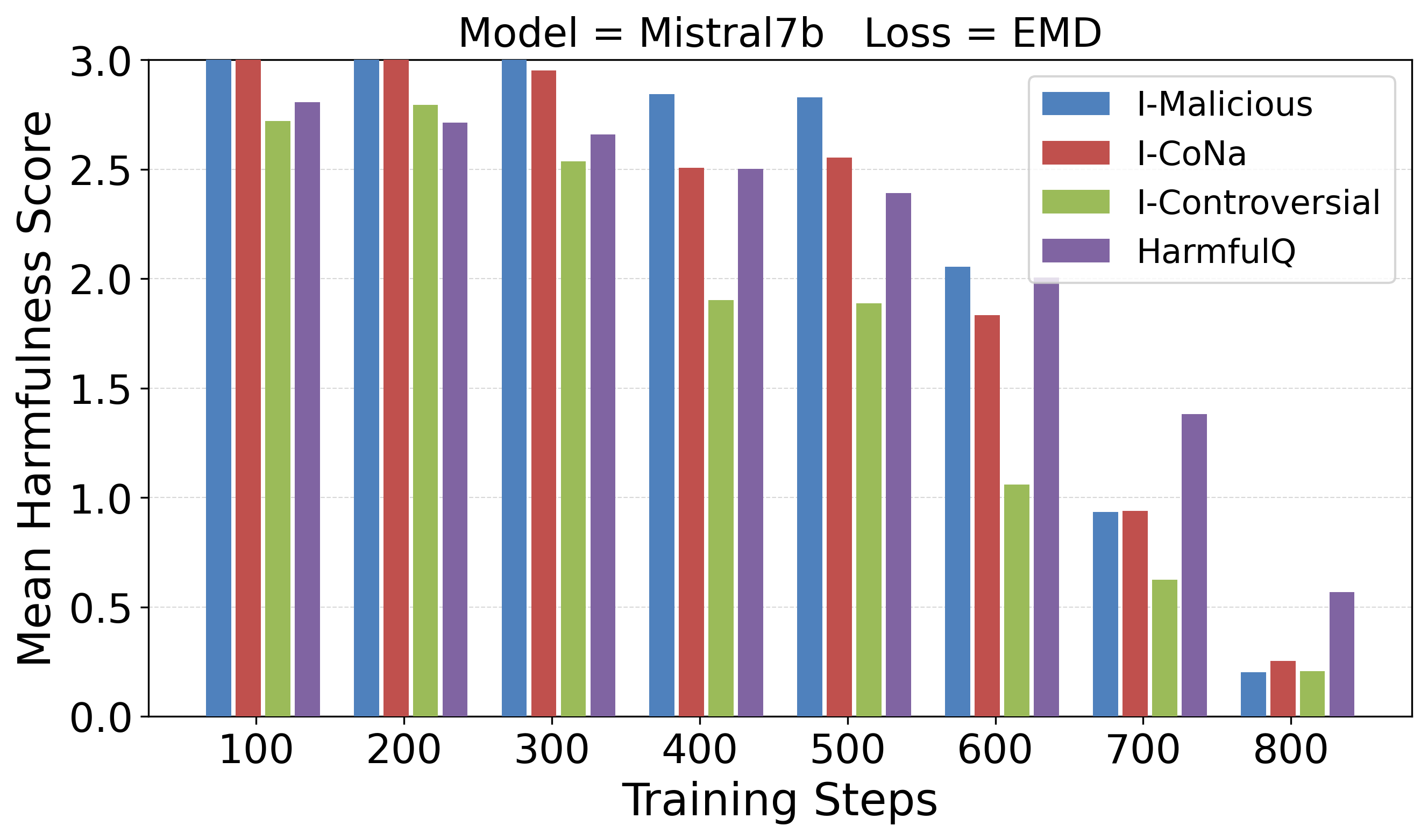}}
	\subfigure[]{
		\includegraphics[width=0.49\linewidth]{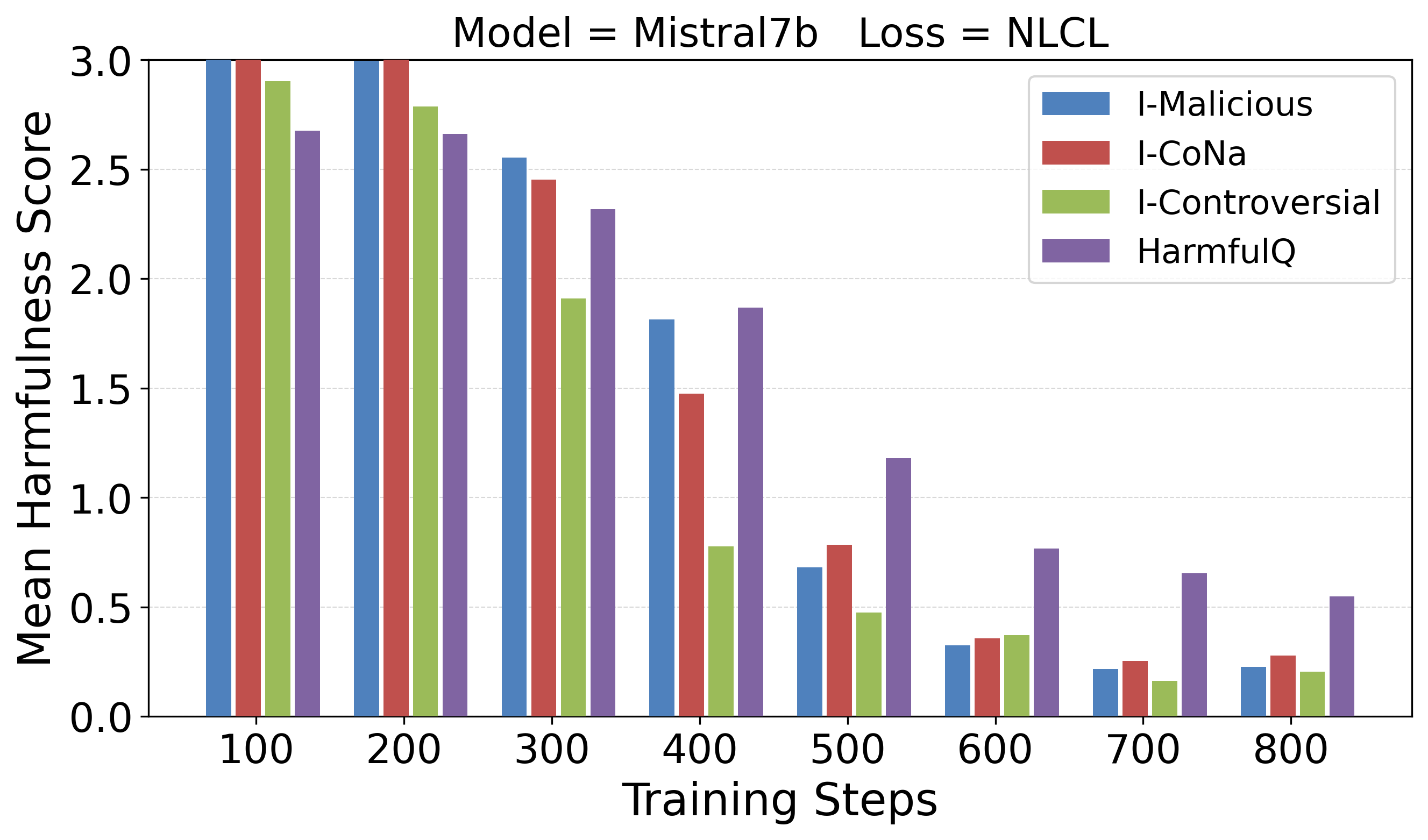}}
	\subfigure[]{
		\includegraphics[width=0.49\linewidth]{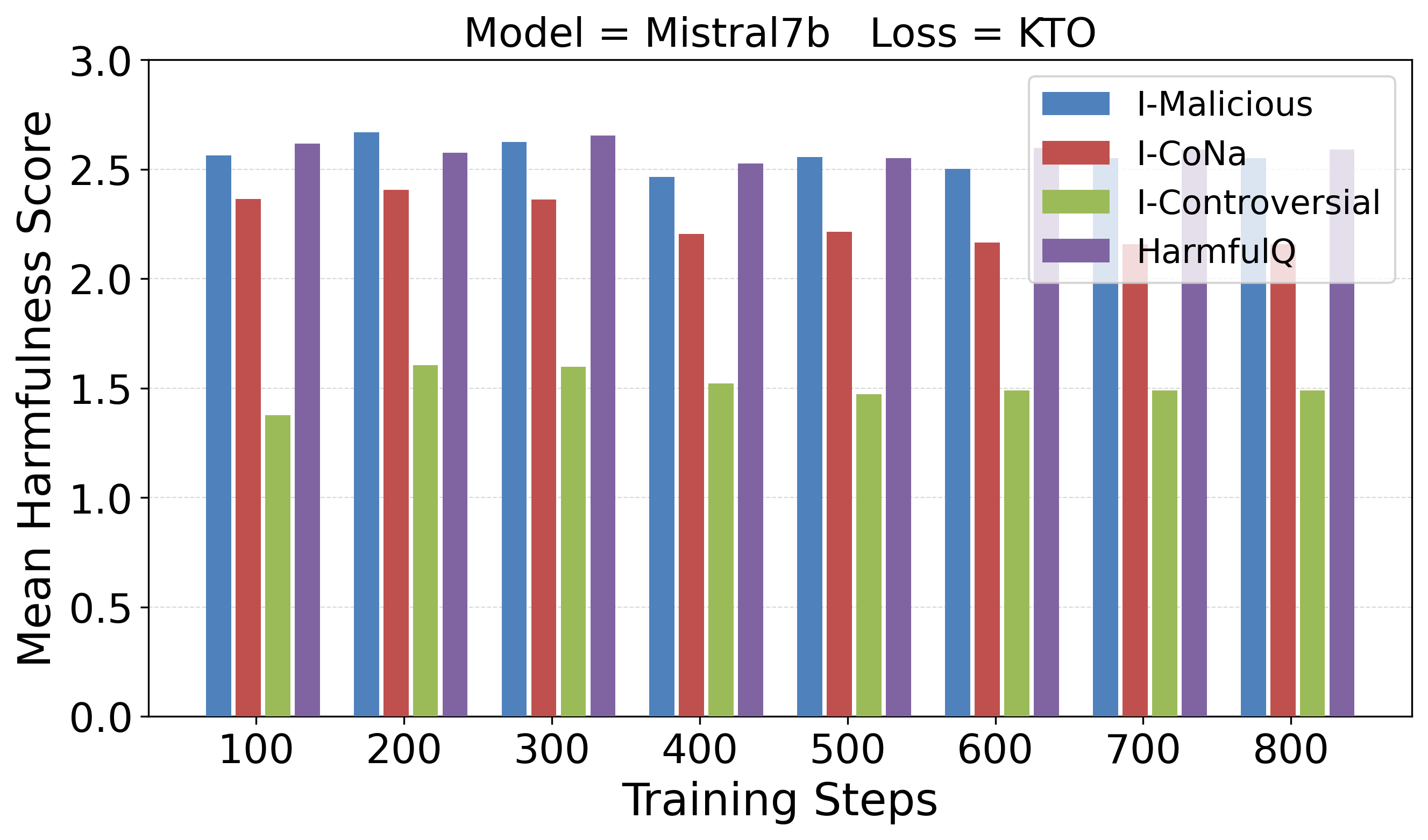}}
	\subfigure[]{
		\includegraphics[width=0.49\linewidth]{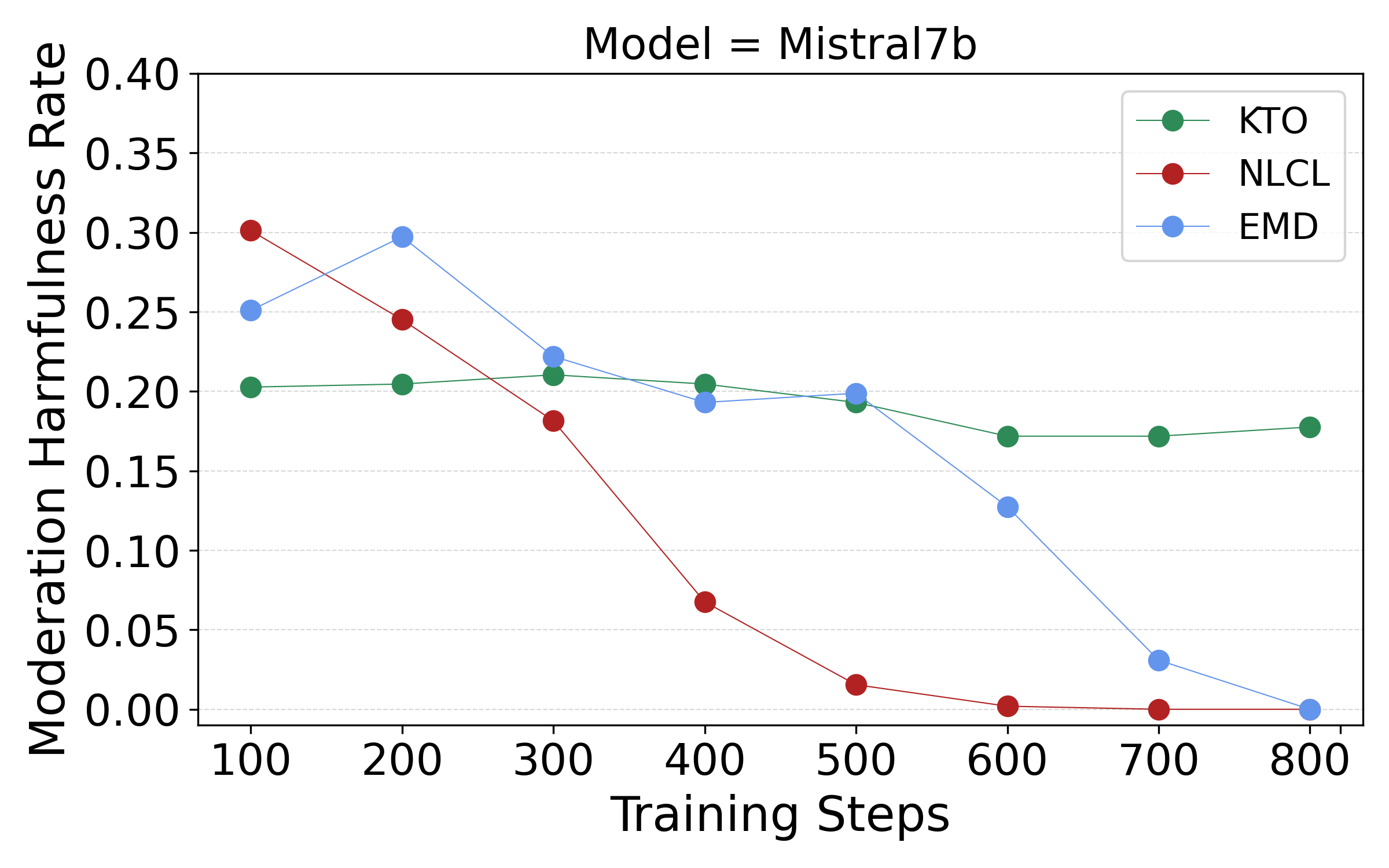}}
	\caption{Response safety evaluation on four harmfulness benchmarks for Mistral 7b. (a)(b)(c) The mean DeBERTa harmfulness score for KTO and our TA-SFT approach with EMD loss and NLCL loss, separately. Lower scores indicate less harmful (safer) responses. (d) The OpenAI Moderation harmful rate. }
	\label{fig:safety_level_harmful_score_mistral7b}
\end{figure}

\begin{figure}[t]
	\centering  
	\subfigure[]{
		\includegraphics[width=0.49\linewidth]{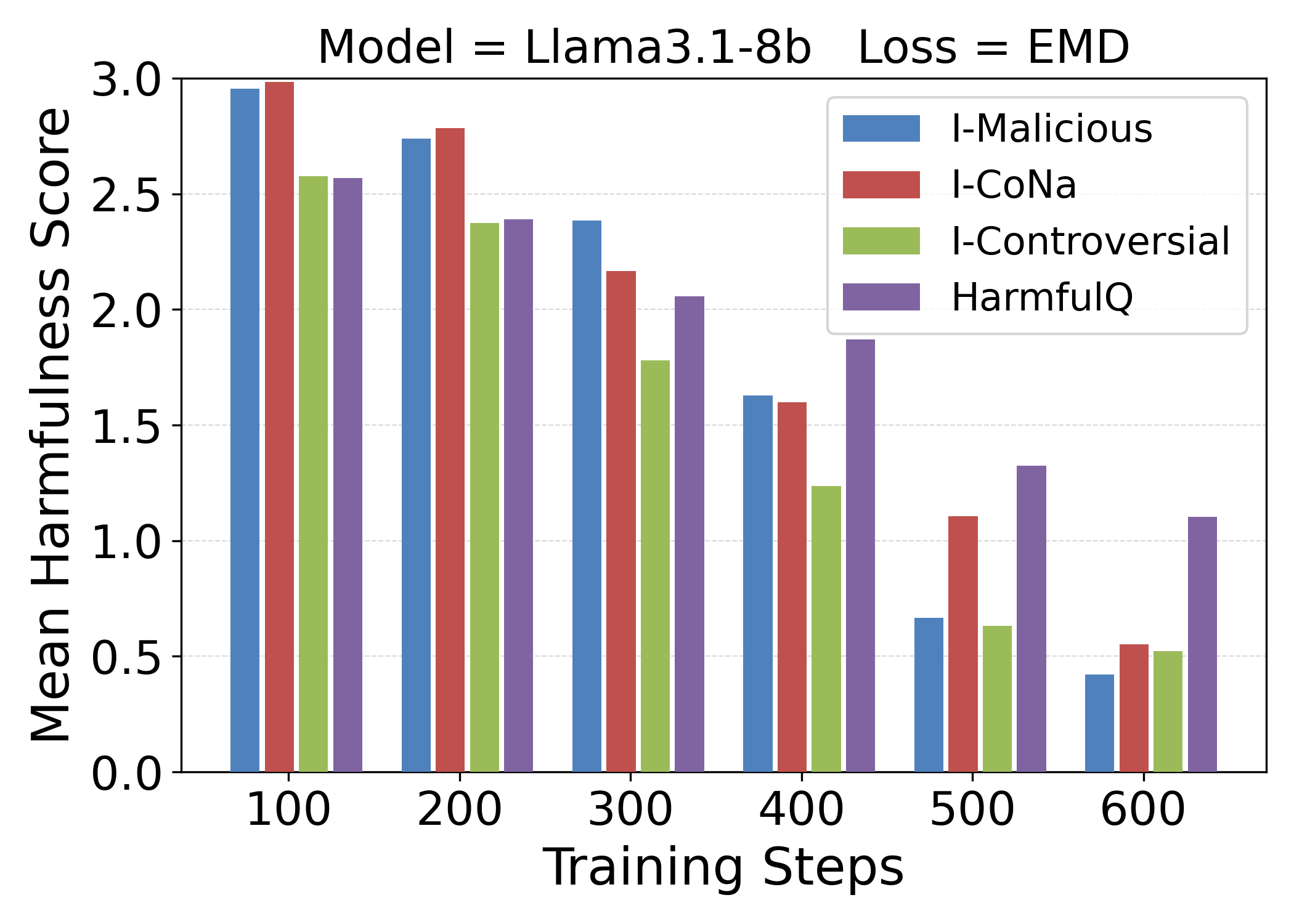}}
	\subfigure[]{
		\includegraphics[width=0.49\linewidth]{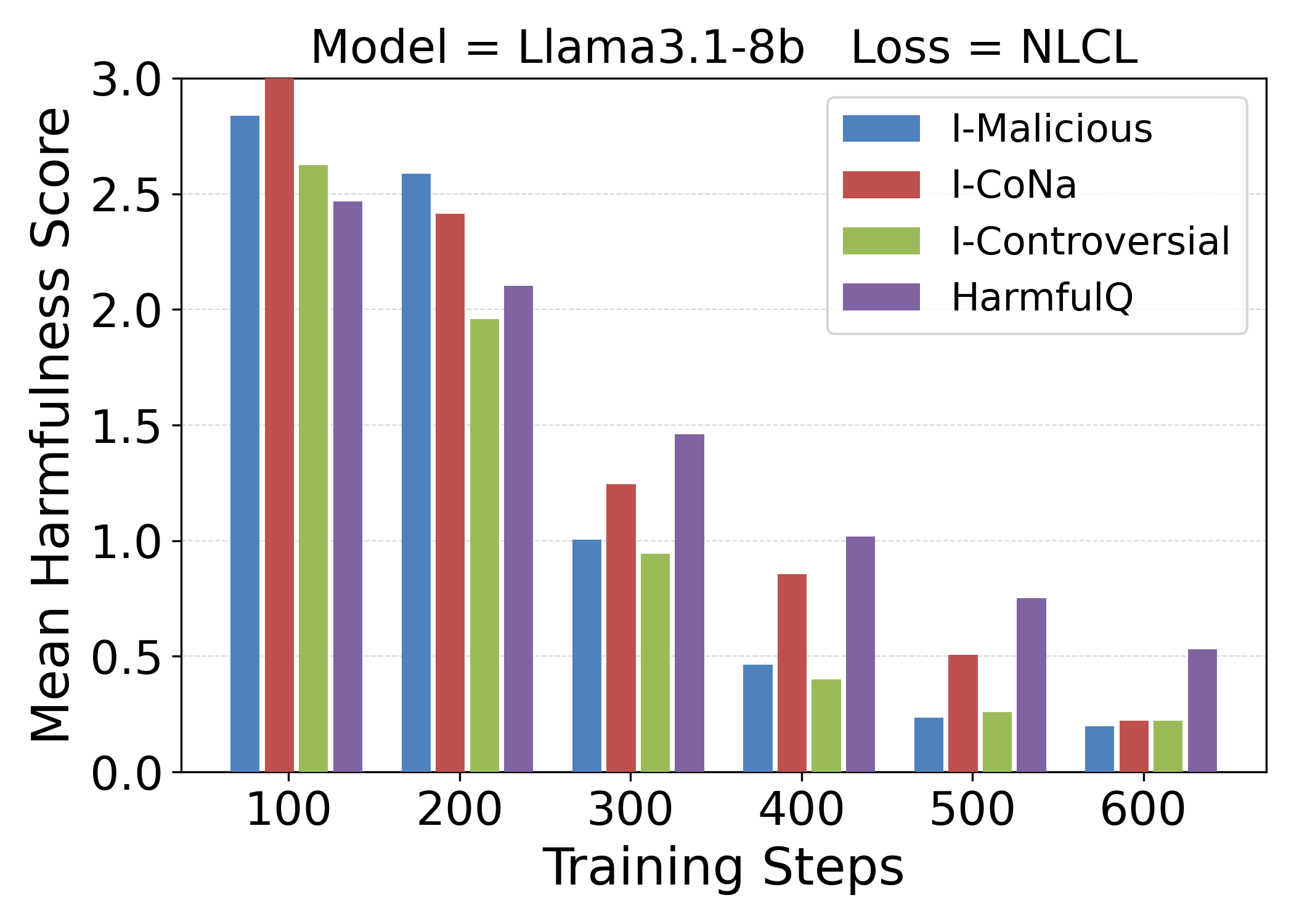}}
	\subfigure[]{
		\includegraphics[width=0.49\linewidth]{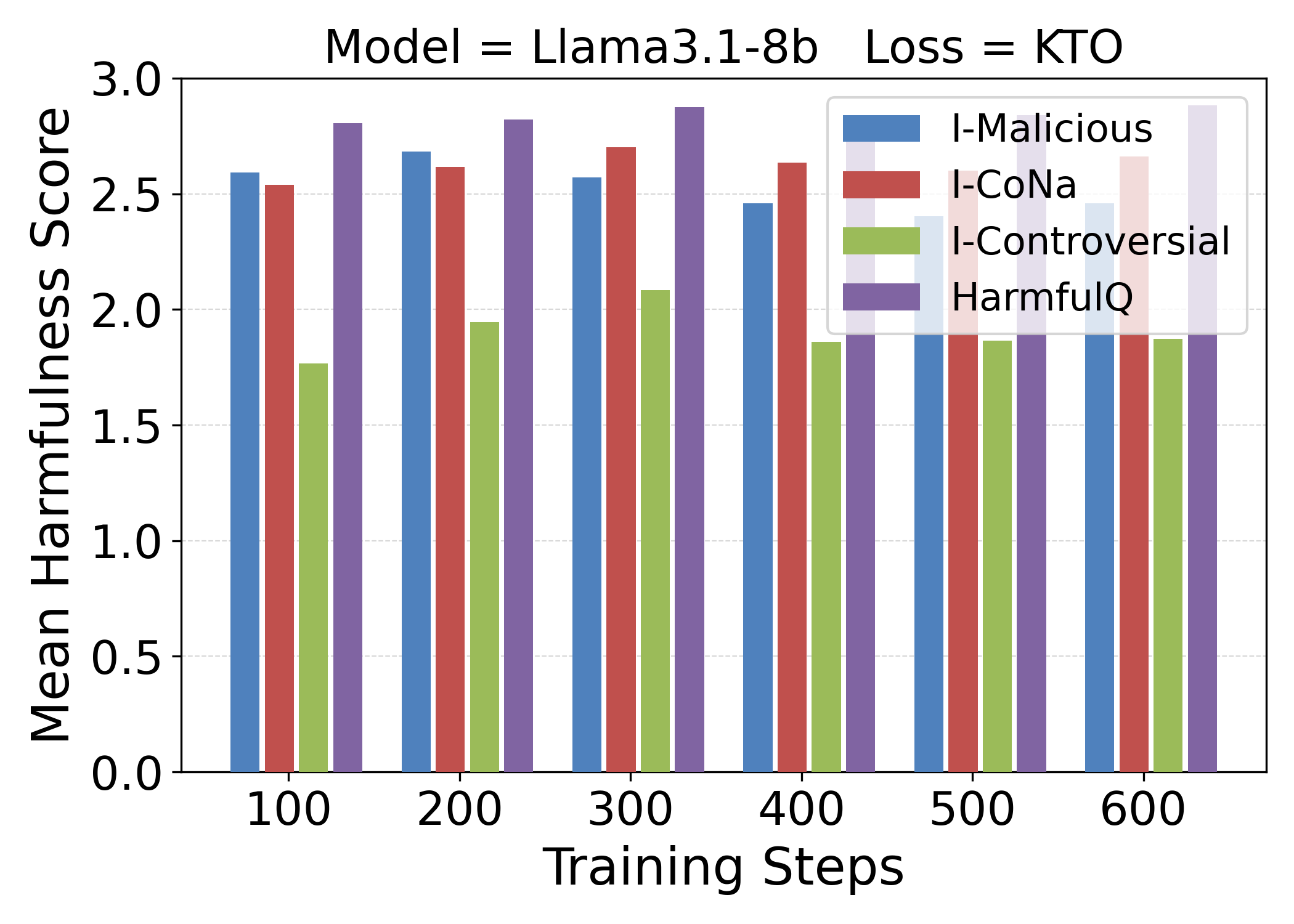}}
	\subfigure[]{
		\includegraphics[width=0.49\linewidth]{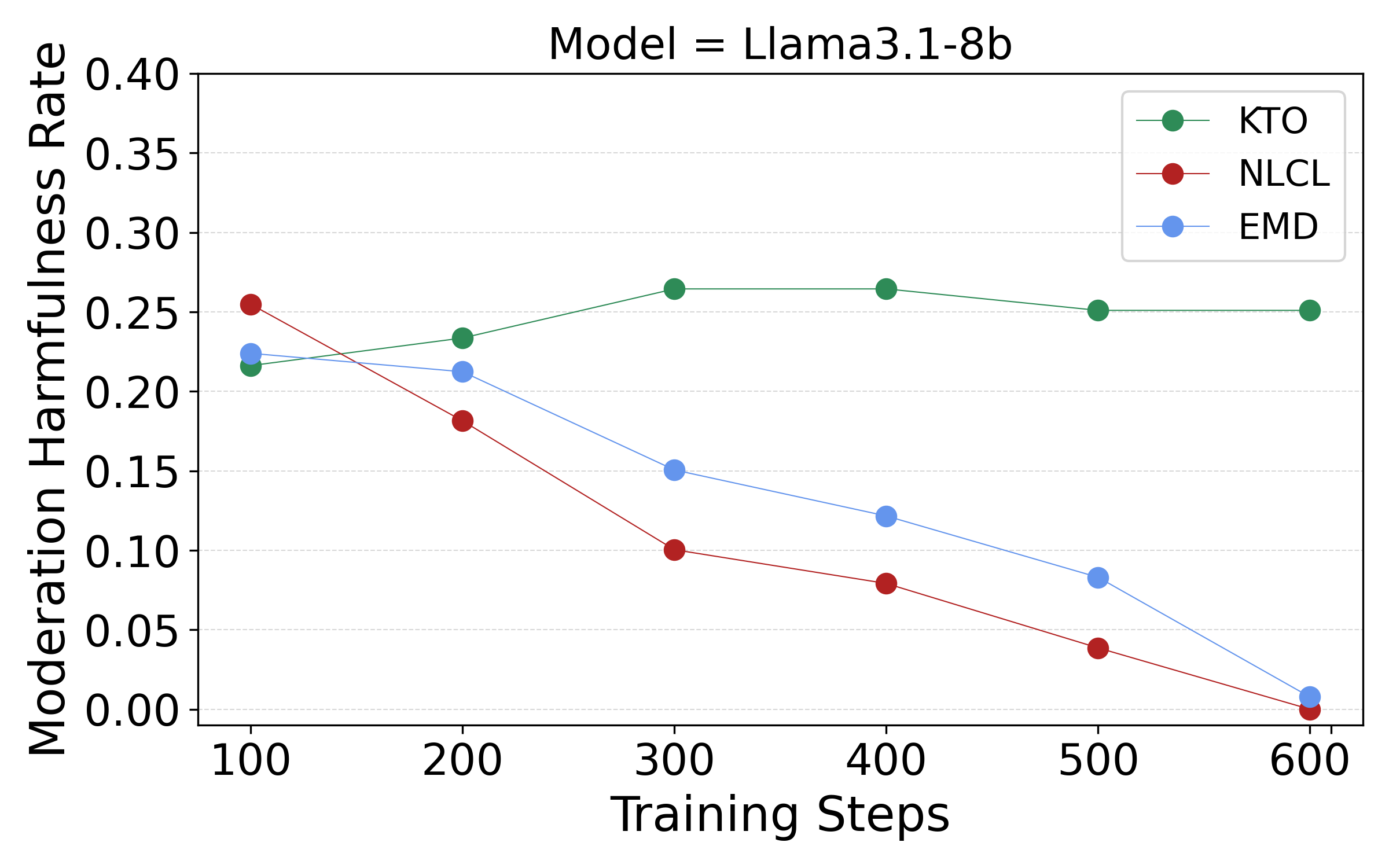}}
	\caption{Response safety evaluation on four harmfulness benchmarks for Llama3.1 8b. (a)(b)(c) The mean DeBERTa harmfulness score for KTO and our TA-SFT approach with EMD loss and NLCL loss, separately. Lower scores indicate less harmful (safer) responses. (d) The OpenAI Moderation harmful rate. }
	\label{fig:safety_level_harmful_score_llama3.1-8b}
\end{figure}

To confirm our results, we also tested our TA-SFT with EMD loss and NLCL loss on Llama 13b (Figure~\ref{fig:safety_level_harmful_score_llama13b}), Mistral 7b (Figure~\ref{fig:safety_level_harmful_score_mistral7b}) and Llama3.1 8b (Figure~\ref{fig:safety_level_harmful_score_llama3.1-8b}). These figures present both the  harmfulness score from DeBERTa model and the harmfulness percentage from OpenAI moderation API. All models exhibit similar to those observed for the Llama 7b model in Section~\ref{sec:safety_level_main} of the main paper, showing a decrease in harmfulness as training progresses using our TA-SFT method, while KTO fails to improve safety levels. Moreover, our TA-SFT approach, with both EMD loss and ORPO loss, ultimately reduces the harmfulness rate to nearly 0\%.

\begin{figure}[t]
	\centering  
	\subfigure[]{
		\includegraphics[width=0.49\linewidth]{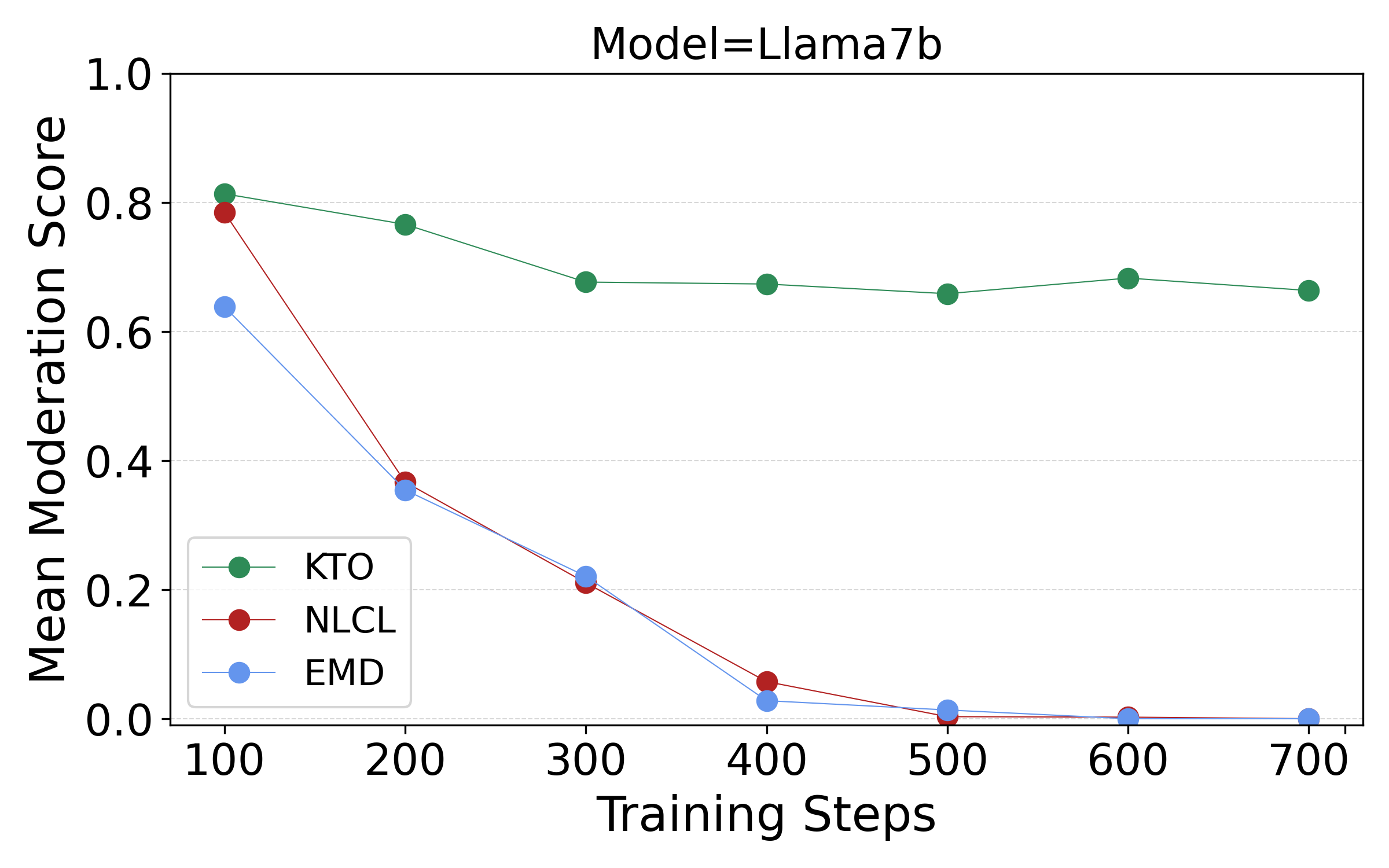}}
	\subfigure[]{
		\includegraphics[width=0.49\linewidth]{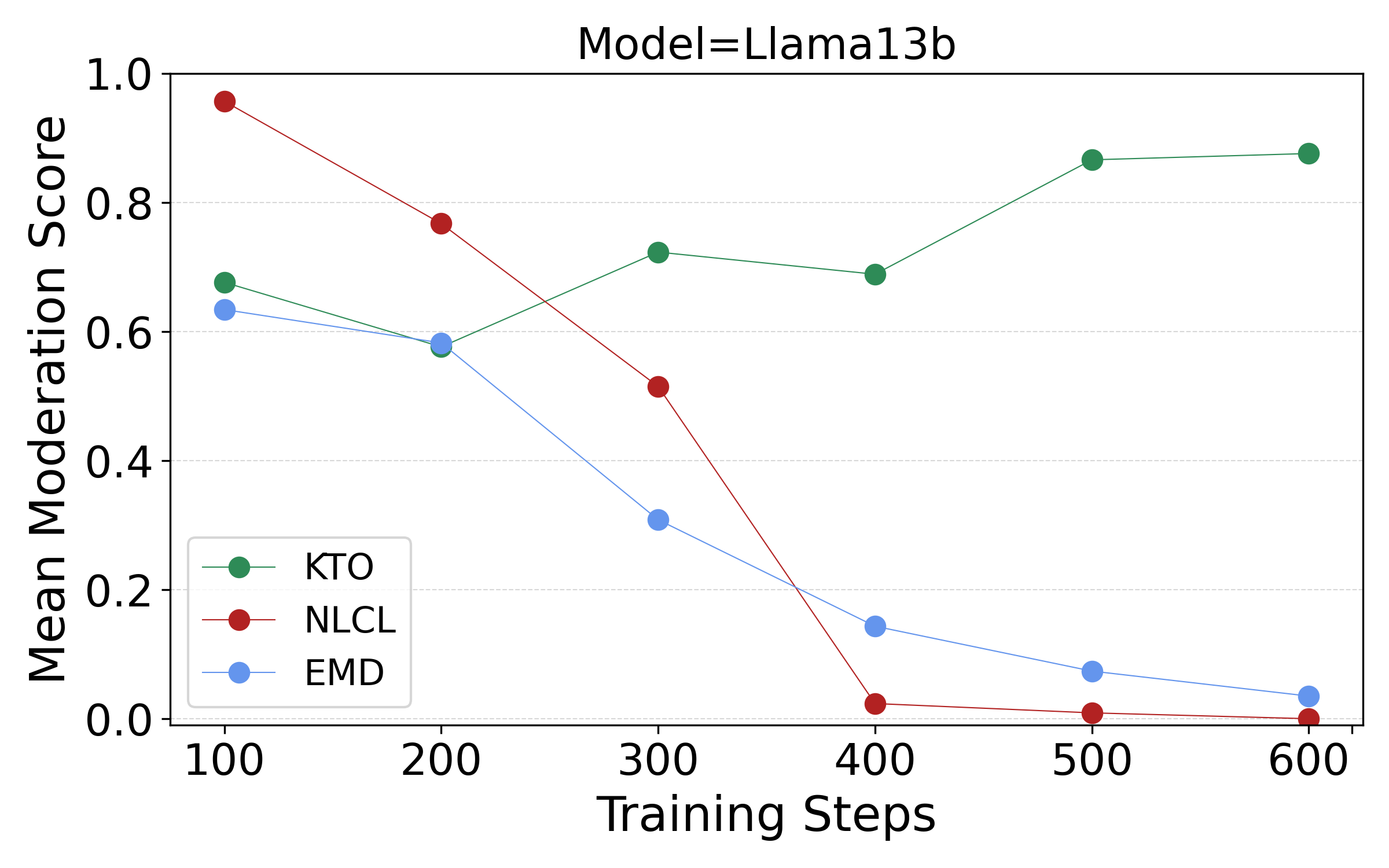}}
	\subfigure[]{
		\includegraphics[width=0.49\linewidth]{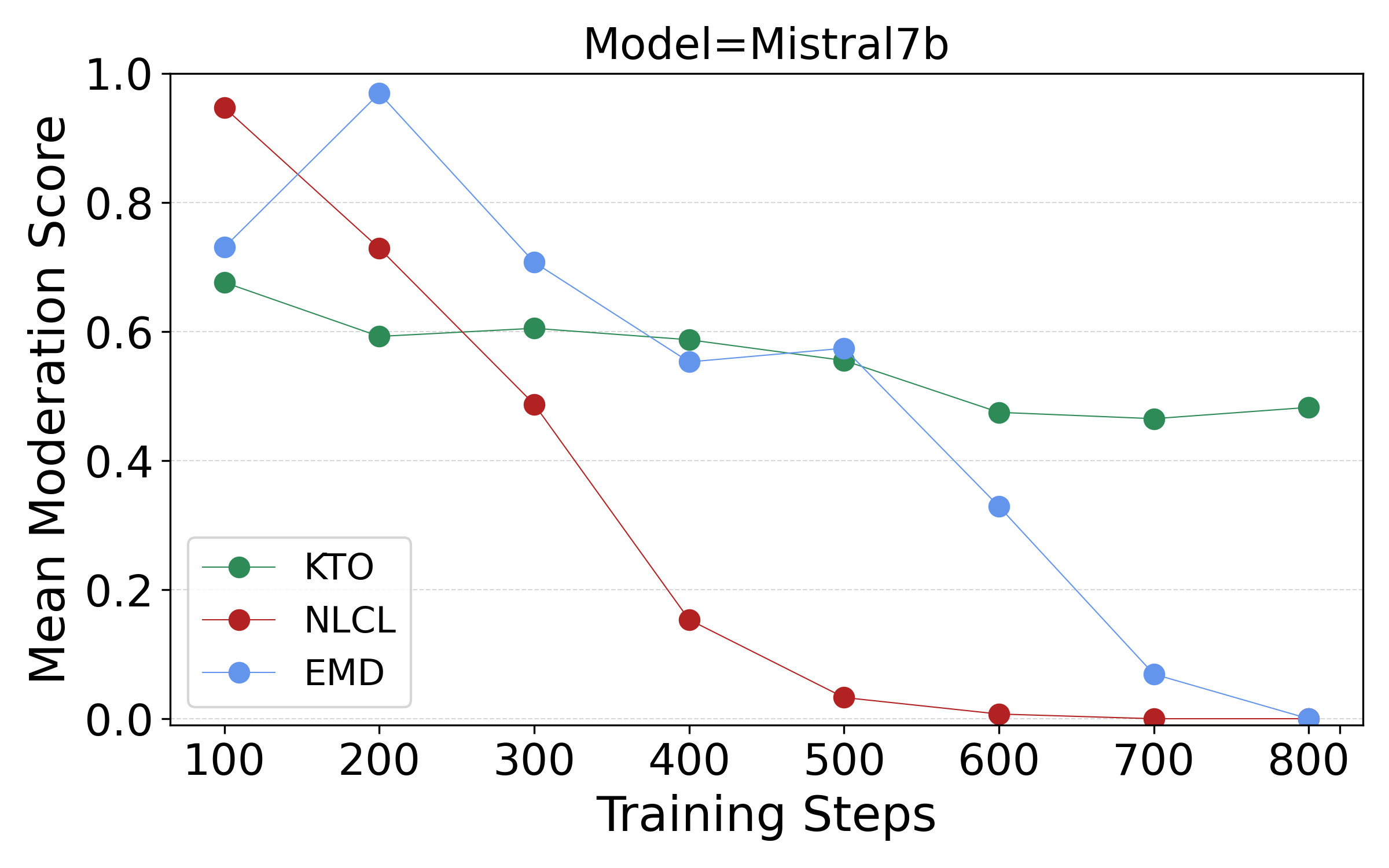}}
	\subfigure[]{
		\includegraphics[width=0.49\linewidth]{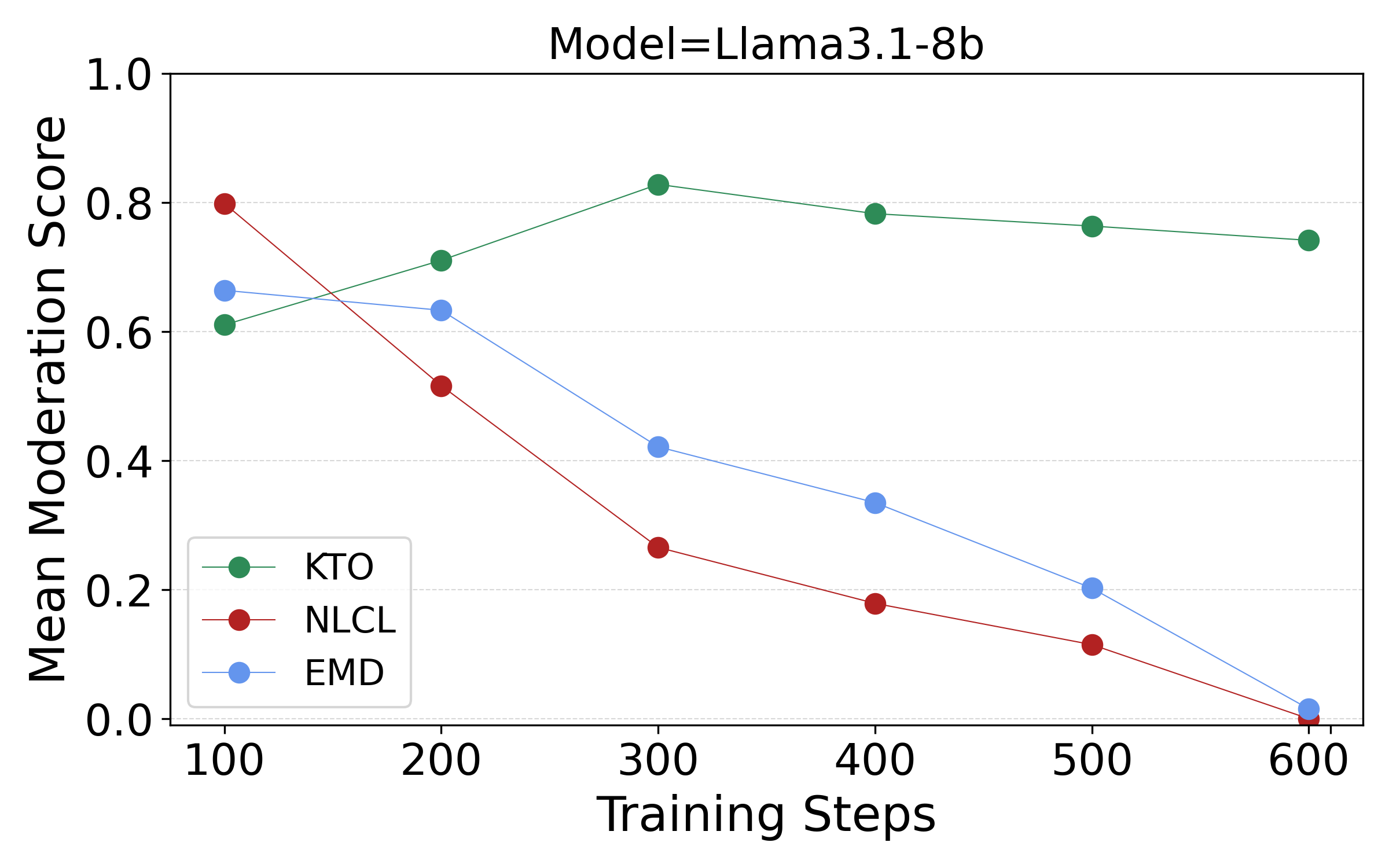}}
	\caption{The averaged OpenAI Moderation harmful scores for KTO and our TA-SFT approach with EMD loss and NLCL loss.}
	\label{fig:moderation_score}
\end{figure}

As stated in the main paper, the OpenAI Moderation API also provide a harmful score beside a binary tag which are shown in Figure~\ref{fig:moderation_score}. The curves in Figure~\ref{fig:moderation_score} representing the average harmfulness score across all responses in the four harmfulness benchmarks, exhibit  a similar trend to the harmfulness rates from the OpenAI Moderation API, depicted in Figure~\ref{fig:safety_level_harmful_score_llama7b}, Figure~\ref{fig:safety_level_harmful_score_llama13b}, Figure~\ref{fig:safety_level_harmful_score_mistral7b}, Figure~\ref{fig:safety_level_harmful_score_llama3.1-8b}, 
\subsubsection{Response Quality}
\label{sec:quality_appendix}
To substantiate the claim that fine tune LLMs with our TA-SFT using both EMD and NLCL loss does not degrade response quality (Section~\ref{sec:quality}), we additionally evaluated the response quality on PIQA and OpenBookQA shown in Table~\ref{table:quality_appendix}.
\begin{table}[!htbp]
\centering
\caption{The response quality of four tested models on additional two multi-choice language modeling benchmarks. There are not
degrading patterns in terms of performance from our TA-SFT approach with EMD loss and NLCL loss.}
\label{table:quality_appendix}
\begin{tabular}{@{}lllllllll@{}}
\toprule
\textbf{}          & \multicolumn{4}{c}{\textbf{PIQA}}                         & \multicolumn{4}{c}{\textbf{OpenBookQA}}                    \\ 
\cmidrule(lr){2-5} \cmidrule(lr){6-9}
\textbf{Model}     & \textbf{SFT} & \textbf{KTO} & \textbf{NLCL} & \textbf{EMD} & \textbf{SFT} & \textbf{KTO} & \textbf{NLCL} & \textbf{EMD} \\ \midrule
\textbf{Llama7b}   & 77.09        & 89.11        & 79.27         & 79.22        & 32           & 35.4         & 35.2          & 34.8         \\
\textbf{Llama13b}  & 75.46        & 79.11        & 79.33         & 79.33        & 35.6         & 34.8         & 34            & 33.4         \\
\textbf{Mistral7b} & 77.31        & 80.85        & 81.23         & 80.85        & 34           & 35.6         & 33.8          & 33.8         \\
\textbf{Llama3.1-8b} & 80.32        & 80.96        & 80.14         & 80.41        & 35           & 37           & 35.2          & 35.2         \\ \bottomrule
\end{tabular}
\end{table}

\subsubsection{Data Efficiency: Fewer Harmful Examples}
\label{sec:Data_Efficiency_appendix}
To confirm the statement that we made in Section~\ref{sec:Data_Efficiency}, we present the number of harmful responses across the four harmfulness benchmarks in Table~\ref{table:fewer_harmuful_examples_appendix}, using our TA-SFT approach with EMD and NLCL. The EMD loss function enables LLMs to learn safe responses with only 100 harmful examples on these two models, whereas the NLCL loss function fails to achieve this.
\begin{table}[t]
\centering
\caption{Number of harmful responses using EMD and NLCL losses with fewer toxic prompts. EMD loss exhibits higher data-efficiency to make LLMs achieve high safety level (lower number of harmful responses) with only 100 toxic prompts in the instruction-tuning dataset.}

\begin{tabular}{@{}lcllllllll@{}}
\toprule
\multicolumn{1}{l}{}                  & \textbf{}           & \multicolumn{2}{c}{\textbf{I-Malicious}}                             & \multicolumn{2}{c}{\textbf{I-CoNa}}                                  & \multicolumn{2}{c}{\textbf{I-Controversial}}                         & \multicolumn{2}{c}{\textbf{HarmfulQ}}                                \\ 
\cmidrule(lr){3-4} \cmidrule(lr){5-6} \cmidrule(lr){7-8} \cmidrule(lr){9-10}
\multicolumn{1}{l}{\textbf{Model}}            & \textbf{\# Toxic} & \multicolumn{1}{c}{\textbf{NLCL}} & \multicolumn{1}{c}{\textbf{EMD}} & \multicolumn{1}{c}{\textbf{NLCL}} & \multicolumn{1}{c}{\textbf{EMD}} & \multicolumn{1}{c}{\textbf{NLCL}} & \multicolumn{1}{c}{\textbf{EMD}} & \multicolumn{1}{c}{\textbf{NLCL}} & \multicolumn{1}{c}{\textbf{EMD}} \\ \midrule
                                      & 1000                & 0                                 & 0                                & 0                                 & 0                                & 0                                 & 0                                & 0                                 & 0                                \\
                                      & 500                 & 1                                 & 1                                & 1                                 & 2                                & 0                                 & 0                                & 0                                 & 0                                \\
                                      & 300                 & 1                                 & 1                                & 8                                 & 5                                & 0                                 & 0                                & 0                                 & 1                                \\
\multirow{-4}{*}{\textbf{Mistral 7b}} & 100                 & 3                                 & 0                                & \cellcolor[HTML]{FD6864}53        & 4                                & 2                                 & 0                                & 4                                 & 1                                \\ \midrule
                                      & 1000                & 0                                 & 0                                & 0                                 & 1                                & 0                                 & 0                                & 0                                 & 3                                \\
                                      & 500                 & 1                                 & 0                                & 6                                 & 6                                & 2                                 & 2                                & 1                                 & 1                                \\
                                      & 300                 & 1                                 & 0                                & 6                                 & 5                                & 0                                 & 2                                & 1                                 & 1                                \\
\multirow{-4}{*}{\textbf{Llama3.1-8b}}  & 100                 & 1                                 & 0                                & \cellcolor[HTML]{FD6864}12        & 7                                & 2                                 & 2                                & 1                                 & 1                                \\ \bottomrule

\end{tabular}
\label{table:fewer_harmuful_examples_appendix}

\end{table}

\subsubsection{Training Data: Safe Samples vs Unsafe Samples}
\label{sec:stl_appendix}
To confirm the observation that we made in Section~\ref{sec:stl}, we compare the performance of Safety-Tuned Llamas (STL) with our TA-SFT approach using EMD loss on Mistral7b and Llama3.1 8b, despite the latter being fine-tuned with a smaller number of harmful data. Although STL benefits from high-quality safe responses to toxic prompts, it is evident that TA-SFT with EMD loss still significantly outperforms STL(Table~\ref{table:fewer_harmuful_examples_STL_appendix}).

\begin{table}[t]
\centering
\caption{Number of harmful responses using EMD and safety-tuned-llamas (STL)~\cite{bianchi2023safety} with fewer toxic prompts. There is a notable increase in the number of harmful responses (indicating a decrease in safety) for STL as the number of safe responses in the instruction-tuning dataset decreases.}
\label{table:fewer_harmuful_examples_STL_appendix}
\begin{tabular}{@{}lcllllllll@{}}
\toprule
\multicolumn{1}{l}{}                  & \textbf{}           & \multicolumn{2}{c}{\textbf{I-Malicious}}                            & \multicolumn{2}{c}{\textbf{I-CoNa}}                                 & \multicolumn{2}{c}{\textbf{I-Controversial}}                        & \multicolumn{2}{c}{\textbf{HarmfulQ}}                               \\ 
\cmidrule(lr){3-4} \cmidrule(lr){5-6} \cmidrule(lr){7-8} \cmidrule(lr){9-10}
\multicolumn{1}{l}{\textbf{Model}}            & \textbf{\# Toxic} & \multicolumn{1}{c}{\textbf{STL}} & \multicolumn{1}{c}{\textbf{EMD}} & \multicolumn{1}{c}{\textbf{STL}} & \multicolumn{1}{c}{\textbf{EMD}} & \multicolumn{1}{c}{\textbf{STL}} & \multicolumn{1}{c}{\textbf{EMD}} & \multicolumn{1}{c}{\textbf{STL}} & \multicolumn{1}{c}{\textbf{EMD}} \\ \midrule
                                      & 1000                & 0                                & 0                                & 0                                & 0                                & 0                                & 0                                & 1                                & 0                                \\
                                      & 500                 & 0                                & 1                                & 1                                & 2                                & 0                                & 0                                & 0                                & 0                                \\
                                      & 300                 & 1                                & 1                                & \cellcolor[HTML]{FD6864}13       & 5                                & 0                                & 0                                & 1                                & 1                                \\
\multirow{-4}{*}{\textbf{Mistral 7b}} & 100                 & 8                                & 0                                & \cellcolor[HTML]{FD6864}64       & 4                                & 1                                & 0                                & 5                                & 1                                \\ \midrule
                                      & 1000                & 0                                & 0                                & 0                                & 1                                & 0                                & 0                                & 1                                & 3                                \\
                                      & 500                 & 1                                & 0                                & 7                                & 6                                & 0                                & 2                                & 1                                & 1                                \\
                                      & 300                 & 2                                & 0                                & \cellcolor[HTML]{FD6864}22       & 5                                & 0                                & 2                                & 3                                & 1                                \\
\multirow{-4}{*}{\textbf{Llama3.1-8b}}  & 100                 & \cellcolor[HTML]{FD6864}11       & 0                                & \cellcolor[HTML]{FD6864}71       & 7                                & 1                                & 2                                & 5                                & 1                                \\ \bottomrule
\end{tabular}
\end{table}

\subsubsection{Over-Alignment of Mistral 7b and Llama3.1 8b}
\begin{figure}[!htbp]
\centering
\begin{minipage}{0.49\textwidth}
\centering
\includegraphics[width=\textwidth]{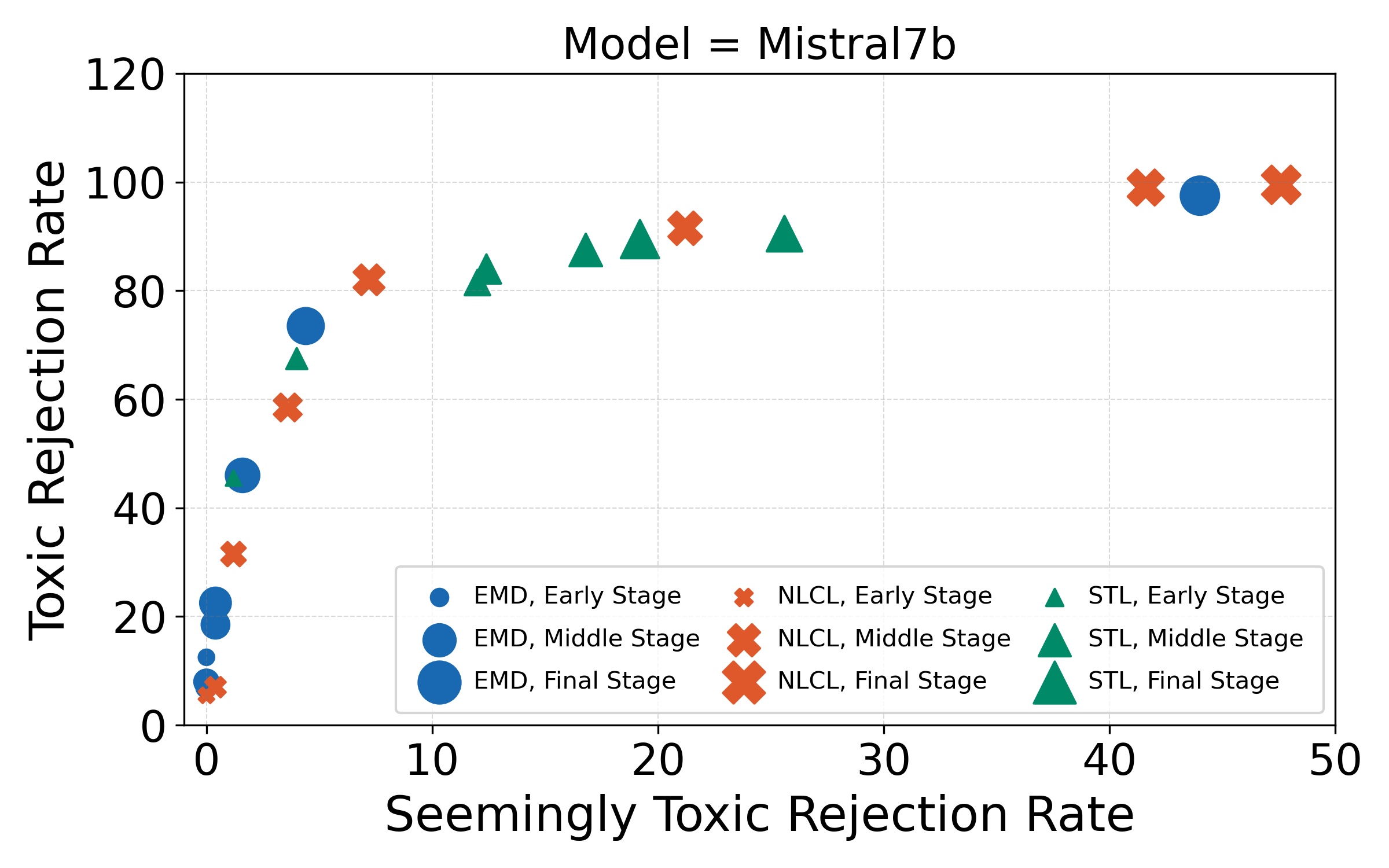}
\end{minipage}
\begin{minipage}{0.49\textwidth}
\centering
\includegraphics[width=\textwidth]{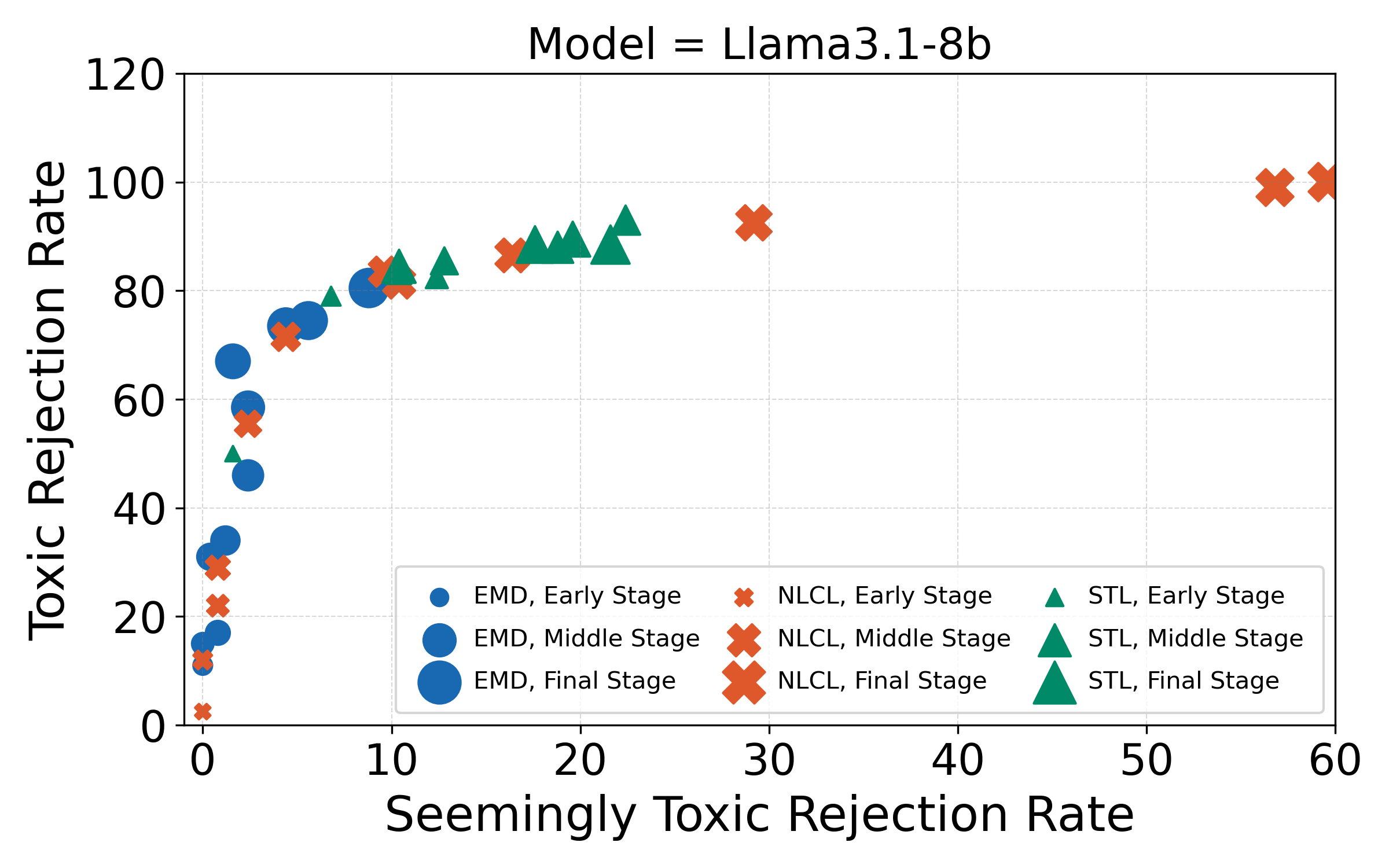}
\end{minipage}
\caption{Over-refusal vs. Safety Levels at different training Stages for Mistral 7b and Llama3.1 8b Models. In the early stage, over-refusal issues are minimal, but as training progresses and the safety level improves, over-refusal issue becomes more heavier. Both TA-SFT and STL show the same trend, empirically demonstrating that the inclusion of refusal examples in the instruction-following dataset is not the cause of the over-refusal issue.}
\label{fig:exaggerated_safety_appendix}
\end{figure}

Consistent with the observation in Section~\ref{sec:Over_Alignment} for Mistral 7b and Llama3.1 8b, as illustrated in Figure~\ref{fig:exaggerated_safety_appendix}, over-refusal issues do not emerge at the beginning of training for Llama 7b and Llama 13b with NLCL and EMD, despite the relatively low safety levels. However, as training progresses, both NLCL and EMD improve the safety of the LLMs but also result in an increased occurrence of over-refusal.
\label{sec:Over_Alignment_appendix}

\begin{table}[]
\centering
\caption{The comparison of Attack Success Rate (ASR) and Mean Judge Score between our method TA-SFT (Ours) and Safety-tuned-llamas (STL).}
\label{table:jailbreaking}
\begin{tabular}{@{}llll@{}}
\toprule
Method & Finetuning Data             & ASR              & Mean Judge Score \\ \midrule
Ours   & Alpaca+1000 unsafe examples & \textbf{19.57\%} & \textbf{6.02}    \\
Ours   & Alpaca+500 unsafe examples  & 26.09\%          & 7.39             \\
Ours   & Alpaca+300 unsafe examples  & 34.78\%          & 8.11             \\
Ours   & Alpaca+100 unsafe examples  & 63.04\%          & 9.13             \\
STL    & Alpaca+1000 safe examples   & 60.00\%          & 8.91             \\
STL    & Alpaca+500 safe examples    & 63.04\%          & 8.71             \\
STL    & Alpaca+300 safe examples    & 73.33\%          & 9.33             \\
STL    & Alpaca+100 safe examples    & 69.57\%          & 9.07             \\ \bottomrule
\end{tabular}
\end{table}

\end{document}